\documentclass[lettersize,journal]{IEEEtran}
\usepackage{amsmath,amsfonts}
\usepackage{algorithmic}
\usepackage{algorithm}
\usepackage{array}
\usepackage[caption=false,font=normalsize,labelfont=sf,textfont=sf]{subfig}
\usepackage{textcomp}
\usepackage{stfloats}
\usepackage{url}
\usepackage{verbatim}
\usepackage{graphicx}
\usepackage{cite}
\usepackage{multirow}
\usepackage{makecell}
\usepackage{epstopdf}
\usepackage{booktabs}
\usepackage{tikz,xcolor}
\usepackage[colorlinks=true,urlcolor=blue,linkcolor=blue,citecolor=blue]{hyperref}
\usepackage[doipre={doi:~}]{uri}

\definecolor{lime}{HTML}{A6CE39}
\DeclareRobustCommand{\orcidicon}{
	\begin{tikzpicture}
		\draw[lime, fill=lime] (0,0)
		circle[radius=0.16]
		node[white]{{\fontfamily{qag}\selectfont \tiny \.{I}D}};
	\end{tikzpicture}
	\hspace{-2mm}
}
\foreach \x in {A, ..., Z}{
	\expandafter\xdef\csname orcid\x\endcsname{\noexpand\href{https://orcid.org/\csname orcidauthor\x\endcsname}{\noexpand\orcidicon}}
}

\begin{document}

\title{A Neural Network Architecture Based on Attention Gate Mechanism for 3D Magnetotelluric Forward Modeling}

\author{
	Xin Zhong\hspace{-1.5mm}\orcidA{}, Weiwei Ling\hspace{-1.5mm}\orcidB{}, Kejia Pan\hspace{-1.5mm}\orcidC{} \textit{Member, IEEE}, Pinxia Wu, Jiajing Zhang, Zhiliang Zhan, Wenbo Xiao 

\thanks{
This work was supported in part by the Research Foundation of the Department of Natural Resources of Hunan Province under Grant HBZ20240154 and Open Fund of SINOPEC Key Laboratory of Geophysics; in part by the Science and Technology Innovation Research Project of Depart
ment of Natural Resources of Jiangxi Province under Grant ZRKJ20232408; and in part by the Jiangxi College of Applied Technology high skill and high-level talents special project (Grant No. JXYY-G2023001).
(\textit{Corresponding author: Weiwei Ling.})

X. Zhong is with the School of information engineering, Jiangxi University of Science and Technology, Ganzhou, 341000, China (e-mail:xszhongxin2@jxyy.edu.cn).

W. Ling, J. Zhang and W. Xiao are with the Jiangxi College of Applied Technology, Ganzhou, 341000, China; W. Ling and J. Zhang also with the Key Laboratory of Ionic Rare Earth Resources and Environment, Ministry of Natural Resources, Ganzhou, 341000, China (e-mail:lingweiwei@jxyy.edu.cn; zhangjj@jxyy.edu.cn; 1807213017@jxyy.edu.cn).

K. Pan is with the Hunan Shaofeng Institute for Applied Mathematics, Xiangtan, 411105, China; and also with the School of Mathematics and Statistics, HNP-LAMA, Central South University, Changsha, 410083, China (e-mail: kejiapan@csu.edu.cn).

P. Wu is with the National Key Laboratory of Computational Physics, Institute of Applied Physics and Computational Mathematics, Beijing, 100088, China (e-mail:wupinxia126@126.com).

Z. Zhan is with the School of science, Jiangxi University of Science and Technology, Ganzhou, Jiangxi, 341000, China (e-mail: 6120240977@mail.jxust.edu.cn).
}}



\maketitle
 
\begin{abstract}
Traditional three-dimensional magnetotelluric (MT) numerical forward modeling methods, such as the finite element method (FEM) and finite volume method (FVM), suffer from high computational costs and low efficiency due to limitations in mesh refinement and computational resources. We propose a novel neural network architecture named MTAGU-Net, which integrates an attention gating mechanism for 3D MT forward modeling. Specifically, a dual-path attention gating module is designed based on forward response data images and embedded in the skip connections between the encoder and decoder. This module enables the fusion of critical anomaly information from shallow feature maps during the decoding of deep feature maps, significantly enhancing the network's capability to extract features from anomalous regions. Furthermore, we introduce a synthetic model generation method utilizing 3D Gaussian random field (GRF), which accurately replicates the electrical structures of real-world geological scenarios with high fidelity. Numerical experiments demonstrate that MTAGU-Net outperforms conventional 3D U-Net in terms of convergence stability and prediction accuracy, with the structural similarity index (SSIM) of the forward response data consistently exceeding 0.98. Moreover, the network can accurately predict forward response data on previously unseen datasets models, demonstrating its strong generalization ability and validating the feasibility and effectiveness of this method in practical applications.
\end{abstract}


\section{Introduction}

 \IEEEPARstart{T}{he} magnetotelluric (MT) method, as a geophysical exploration technique, has become one of the mainstream electromagnetic exploration methods due to its low cost, ease of operation, and large exploration depth range~\cite{ref1,ref2}. It is widely applied in various fields, including mineral resource exploration, electrical structure research, oil and gas energy exploration, and deep earth structural studies~\cite{ref3,ref4,ref5,ref6,ref7,ref8}. The forward modeling is the prerequisite and foundation for inversion, and its computational speed and accuracy directly affect the performance of inversion~\cite{ref9}. 
 
 Conventional numerical approximation methods for forward modeling mainly include the finite difference method (FDM), finite volume method (FVM), and finite element method (FEM)~\cite{ref10,ref11,ref12}. The main idea of these methods is to discretize the geoelectric model by dividing the large domain into grids, and then solve the problem approximately with the help of computers~\cite{ref13}. For instance, FDM discretizes differential equations by converting derivative terms into finite-difference approximations on structured orthogonal grids. While FDM is simple and computationally efficient, it struggles with complex topography or heterogeneous anomalies~\cite{ref14}. FVM constructs discrete equations through control-volume integration, ensuring local conservation but suffering from reduced computational efficiency due to demanding grid generation and integration processes~\cite{ref15}. In contrast, FEM employs unstructured meshes, enabling flexible adaptation to arbitrary geometries. By achieving high-resolution discretization of localized regions, FEM accurately solves electromagnetic field problems in complex geological models~\cite{ref16,ref17,ref18}.
 
 Although these traditional numerical forward modeling methods can simulate the subsurface electrical structures with high accuracy, they are limited by grid discretization and hardware resources. Moreover, as the dimensionality increases, the computational complexity grows exponentially\cite{ref19}. In one-dimensional (1D) and two-dimensional (2D) scenarios, the computational load is relatively manageable. However, in three-dimensional (3D) scenarios, due to the complexity and variability of the model’s anomalous spatial position information, the forward modeling response correlations between different anomalies extend beyond the 2D plane. This significantly increases the degrees of freedom and results in a dramatic rise in computational cost~\cite{ref20,ref21,ref22,ref23}. Therefore, there is an urgent need to develop more efficient 3D forward modeling methods to improve computational efficiency and meet the demands of large-scale and complex geological model simulations.
 
 In recent years, the use of deep learning to solve forward and inversion problems in electromagnetic fields has attracted significant attention from researchers and has been successfully applied in several studies~\cite{ref24,ref25,ref26,ref27,ref28,ref29}. For example, Liao Weiyang et al. developed an efficient MT forward simulation network (EFDO) using an extended Fourier deep architecture for 2D MT forward simulation in complex continuous media~\cite{ref30}. Peng Zhong et al. proposed a method for solving electromagnetic forward problems based on a Fourier neural network operator, which is at least 100 times faster than the conventional finite difference method~\cite{ref31}. Deng Fei et al. constructed a network structure based on convolutional bidirectional long short-term memory (Conv-BiLSTM) and extended convolution (D-LinkNet) for 2D magnetotelluric forward modeling~\cite{ref32}. Lim, Joowon et al. proposed a novel approach to train a deep neural network by using the residuals of Maxwell’s equations as the physical-driven loss function for the network, enabling fast and accurate solutions of Maxwell’s equations without relying on other computational electromagnetic solvers~\cite{ref33}. Tao Shan et al. developed a dual-branch convolutional network for 2D magnetotelluric forward modeling, predicting the apparent resistivity and impedance phase of the electromagnetic model~\cite{ref34}. Wang Xuben et al. constructed a Transformer-based network structure capable of predicting the apparent resistivity and phase at different polarization directions, achieving deep learning forward modeling \cite{ref35}.
 
The aforementioned research has implemented deep learning-based MT forward modeling, but most studies are still focused on 1D and 2D scenarios, with relatively few studies on 3D forward modeling. Furthermore, the datasets constructed for 3D forward modeling are mostly regular anomalous bodies, making it difficult to simulate the actual complex geological environment~\cite{ref35}. Considering the complex and diverse electrical structures of real geological formations, studying 3D forward modeling for complex geoelectric models is one of the urgent problems to be addressed.

In this paper, we apply deep learning techniques to the 3D MT forward modeling problem. By designing a neural network architecture based on dual-path attention gate mechanism, we achieve an end-to-end mapping from input (resistivity model parameters) to output (response data). Furthermore, to simulate resistivity models with complex terrains, we introduce 3D Gaussian random field (GRF) to generate complex, continuous resistivity models controlled by spectral methods, creating a dataset of 3D model samples. Finally, we conduct numerical experiments and comparative analysis on the constructed network architecture to evaluate the feasibility and effectiveness of this method in practical applications.

\section{Methodology}\label{II}
\subsection{Finite Element 3D Forward Modeling}
\label{subA}
In the frequency domain electromagnetic field, the time harmonic factor $e^{\mathrm{i} \omega t}$ is used, and it satisfies the following equations:
\begin{equation}
	\label{eq1}
	\begin{aligned}
		& \nabla \times \mathbf{E}=-\mathrm{i}\omega \mu \mathbf{H}, \\
		& \nabla \times \mathbf{H}=\mathbf{J}+\sigma \mathbf{E}.
	\end{aligned}
\end{equation}
Where, $\omega$ is the angular frequency, $\mu$ is the vacuum magnetic permeability, $\sigma$ is the electric conductivity, $\mathbf{J}$ is the source term, $\mathbf{E}=({E}_{x},{E}_{y},{E}_{z})$ denotes the electric field and $\mathbf{H}=({H}_{x}, {H}_{y}, {H}_{z})$ represents the magnetic field.

By substituting the electric field equation into the magnetic field equation, the following equation is obtained:
\begin{equation}
	\label{eq2}
	\nabla \times \nabla \times \mathbf{E}+\mathrm{i} \omega \mu \sigma \mathbf{E}=-\mathrm{i} \omega \mu \sigma \mathbf{J}.
\end{equation}

\begin{figure}[!htp]
	\centering
	\includegraphics[width=3.5in]{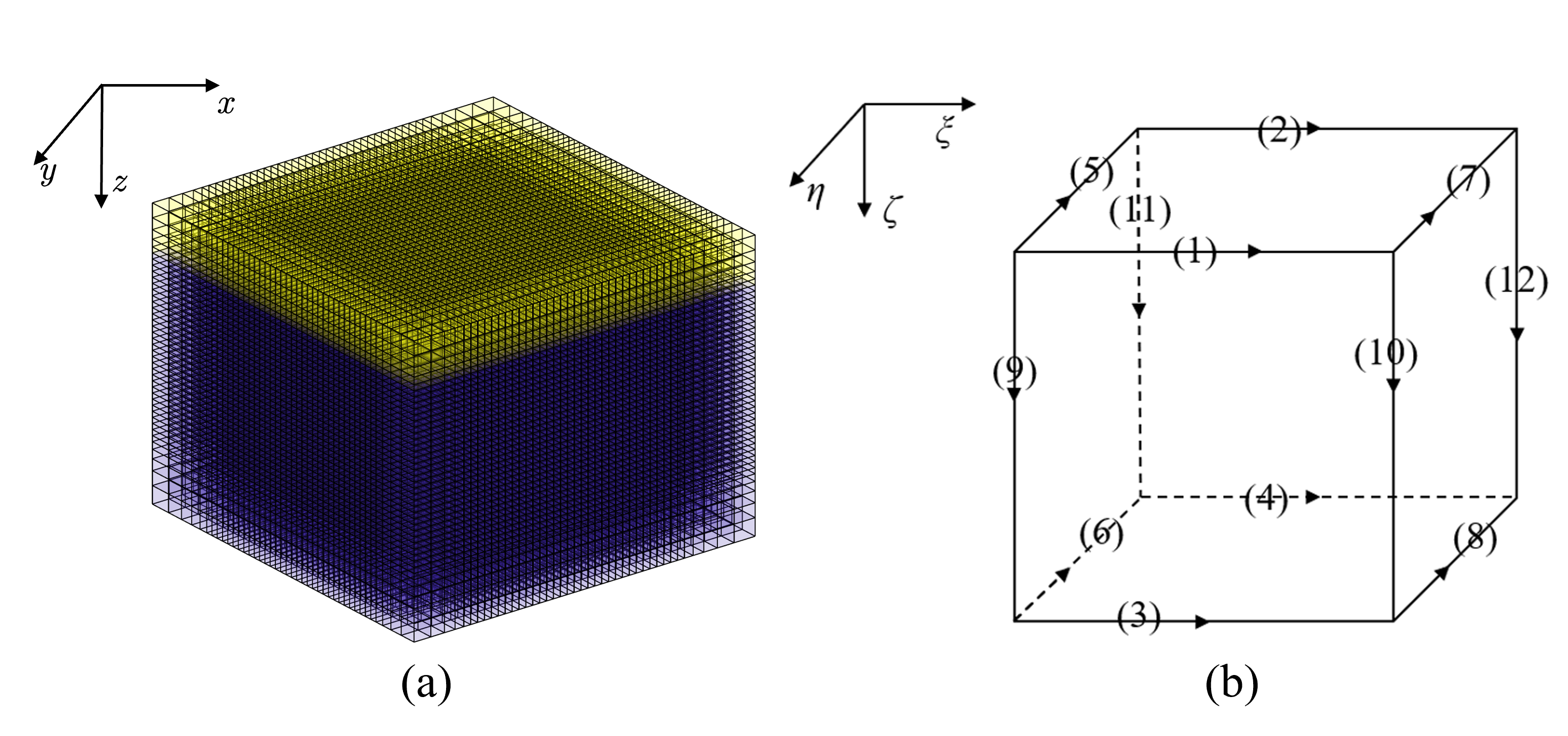}
	\caption{Schematic diagram of 3D computational domain grid and hexahedral element. (a) is 3D computational domain grid, and (b) is the hexahedral element}.
	\label{fig1_1}
\end{figure}

The research domain for the 3D magnetotelluric forward problem is shown in Fig. \ref{fig1_1}(a), with the solution domain is a regular hexahedron that includes an air layer (the yellow area). The hexahedral elements shown in Fig. \ref{fig1_1}(b) are used to discretize equation \eqref{eq2}, ensuring the continuity of the electric field. The vector basis function $\mathbf{N}$ is defined on each edge and belongs to  $H({\bf \textrm{curl}}; \Omega)$ (the space of vector functions whose curl is square-integrable). By taking the dot product of both sides with the test function $\mathbf{V}$ for equation (\ref{eq2}) and integrating over the entire region, we obtain:

\begin{equation}
	\label{eq3}
	\iiint_{\Omega}(\nabla \times \nabla \times \mathbf{E}+\mathrm{i} \omega \mu \sigma \mathbf{E})\cdot \mathbf{V}d \Omega =-\mathrm{i} \omega \mu \iiint_{\Omega}\mathbf{J}\cdot \mathbf{V}d\Omega.
\end{equation}

Using vector identities and applying integration by parts, equation \eqref{eq3} can be rewritten as:
\begin{equation}
	\label{eq4}
	\iiint_{\Omega}(\nabla \times \mathbf{E} \cdot \nabla \times \mathbf{V}+\mathrm{i} \omega \mu \sigma \mathbf{E}\cdot\mathbf{V})d \Omega =-\mathrm{i} \omega \mu \iiint_{\Omega}\mathbf{J}\cdot \mathbf{V}d\Omega.
\end{equation}

For magnetotelluric, there is no electromagnetic source, $\mathbf{J}$ can be set to be zero. The following boundary condition (Dirichlet boundary) can be applied to the boundaries:
\begin{equation}
	\label{eq5}
	\mathbf{n}\times\mathbf{E}=\mathbf{n} \times \mathbf{E}_{0},
\end{equation}
where $\mathbf{n}$ is the unit outer normal direction, $\mathbf{E}_{0}$ is the one-dimensional electromagnetic response of the medium on the boundary, which can be obtained by analytical methods.

According to the derivation process described in reference \cite{ref22}, the impedance tensor $\mathbf{Z}$ can be obtained by converting the electromagnetic field components ($E_{x}^{(1)}, E_{y}^{(1)}, H_{x}^{(1)}, H_{y}^{(1)}$) and ($E_{x}^{(\text{2})}, E_{y}^{(\text{2})}, H_{x}^{(\text{2})}, H_{y}^{(\text{2})}$) at the ground surface with two different polarization directions:
\begin{equation}
	\label{eq13}
	\begin{aligned}
		& {{Z}_{xx}}=(E_{x}^{(1)}H_{y}^{(2)}-E_{x}^{(2)}H_{y}^{(1)})/\zeta , \\
		& {{Z}_{xy}}=(E_{x}^{(2)}H_{x}^{(1)}-E_{x}^{(1)}H_{x}^{(2)})/\zeta , \\
		& {{Z}_{yx}}=(E_{y}^{(1)}H_{y}^{(2)}-E_{y}^{(2)}H_{y}^{(1)})/\zeta , \\
		& {{Z}_{yy}}=(E_{y}^{(2)}H_{x}^{(1)}-E_{y}^{(1)}H_{x}^{(2)})/\zeta ,\\
	\end{aligned}
\end{equation}
where $\zeta=H_y^{(2)}H_x^{(1)}-H_x^{(2)}H_y^{(1)}$,  and ${{Z}_{{xx}}},{{Z}_{{xy}}}, {{Z}_{{yx}}},{{Z}_{zz}}$ are the four components of the impedance tensor $\mathbf{Z}$.

Further, we can get the apparent resistivity and phase calculation equations:
\begin{equation}
	\label{eq14}
	\rho_{ij}= \frac{1}{\mu \omega}|Z_{ij}|^{2},\quad \phi _{ij}= \arctan \frac{\textrm{Im}(Z_{ij})}{\textrm{Re}(Z_{ij})},
\end{equation}
where $i$ and $j$ can respectively be $x$ or $y$.

\subsection{Deep Learning 3D Forward Modeling}
The core idea of deep learning-based forward modeling is to train a neural network model to approximate traditional forward operators, thereby realizing the mapping relationship between the 3D resistivity model and the forward response data. Specifically, the goal of forward simulation is to compute the observed data (such as apparent resistivity and phase) for a given resistivity model, which can be defined as:
\begin{equation}
\label{eqa1}
\hat{d}=F(m, \theta)
\end{equation}
Here, $F$ is the forward operator of the neural network, $\hat{d}$ is the predicted forward response data, $m$ is the resistivity model parameters, and $\theta$ represents the weights and biases of the network.

\subsection{Neural Network Architecture}
\begin{figure*}[!htp]
	\centering
	\includegraphics[width=5.5in]{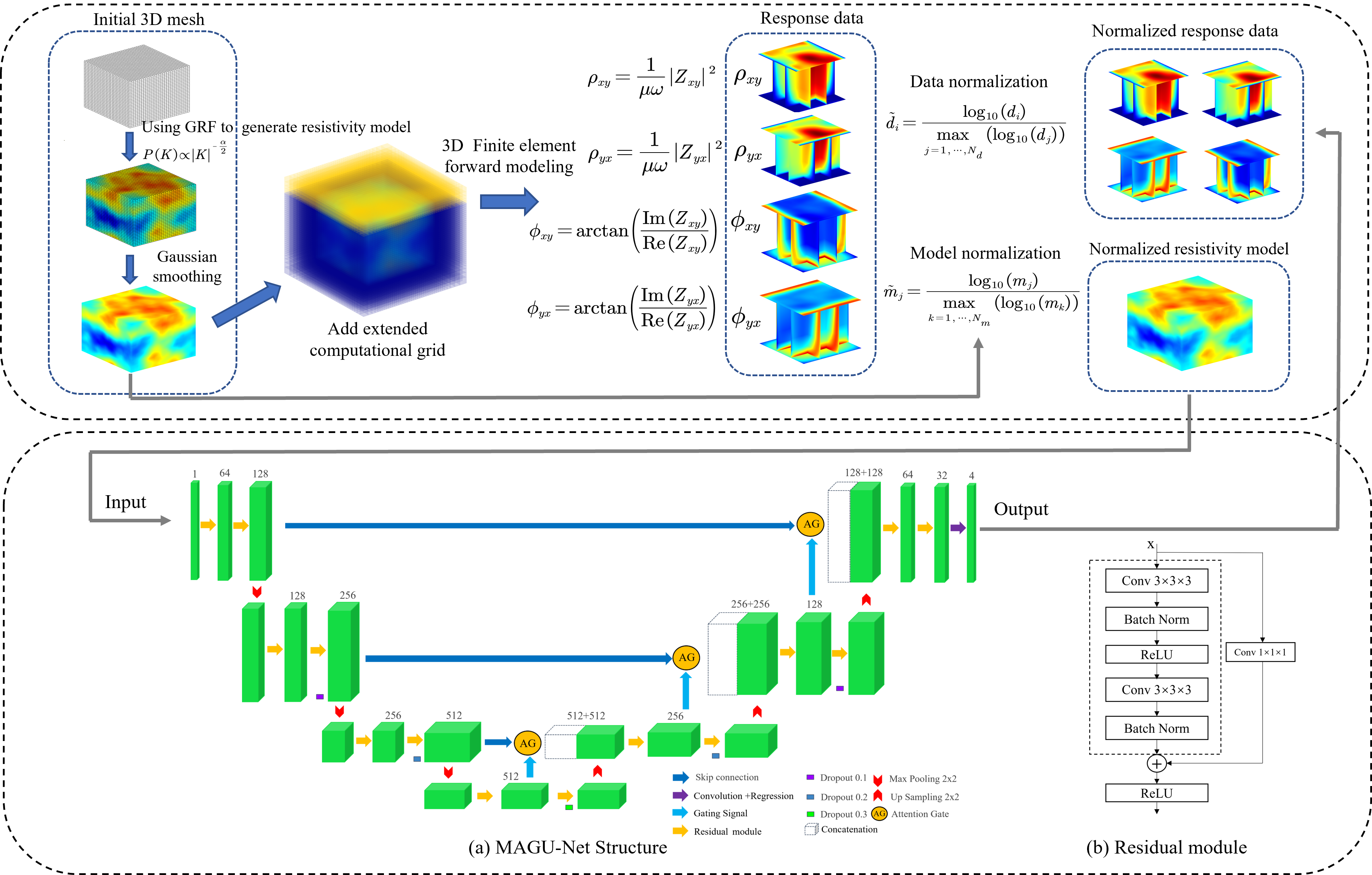}
	\caption{MTAGU-Net Attention gate network architecture diagram.}
	\label{fig1}
\end{figure*}

In this paper, we propose a three-dimensional neural network architecture named MTAGU-Net (as shown in Fig. \ref{fig1}) incorporating dual-path attention gating mechanisms. The network features a three-layer symmetric encoder-decoder architecture. During the encoding phase, multi-level features are extracted through cascaded down-sampling. Each encoder module contains a residual convolutional block and a $2\times 2\times 2$ max-pooling layer. The shallow encoder module outputs high-resolution feature maps (1/2 the input size), preserving edge details and spatial location information of resistivity anomalies. The intermediate encoder module captures local resistivity variation patterns through secondary down-sampling (1/4 the input size). The deep encoder module generates low-resolution feature maps (1/8 the input size) after tertiary down-sampling, encapsulating global contextual semantic information. In the decoder stage, progressive up-sampling is used to restore spatial resolution. Each decoder module first employs 3D transposed convolution to double the size of the feature maps, followed by cross-layer fusion with high-resolution features from the corresponding encoder layer ~\cite{ref35,ref36,ref37,ref38}. 
To address the vanishing gradient problem during deep network training, residual modules are designed in both the encoder and decoder to increase the network's depth, thereby enhancing its ability to extract deep features. All residual blocks adopt two sets of "Conv3D-BN-ReLU" units, with identity mapping preserving the original gradient path to mitigate the vanishing gradient issue~\cite{ref39}. Additionally, an adaptive channel dropout mechanism (dynamically adjusted between 0.1 and 0.3) is introduced after each down-sampling or up-sampling operation. This strategy, tailored to the characteristics of 3D model data, randomly masks feature maps at the channel level, reducing the risk of overfitting while maximally preserving the spatial continuity of anomaly structures~\cite{ref40}.

Finally, skip connections between corresponding encoder-decoder layers propagate high-resolution feature maps (containing spatial localization information) from the encoder to the decoder, where they are fused with low-resolution feature maps (embodying contextual semantic information) to capture multi-scale features. However, while this approach enables hierarchical feature integration, it suffers from an inherent limitation: the isotropic treatment of intermediate semantic influences, where all spatial regions in the input resistivity model are assigned equal importance weights (e.g., failing to distinguish between resistivity anomaly regions and boundary zones)~\cite{ref41}. Consequently, the network lacks adaptive focus or weighting mechanisms for critical regions, particularly under complex 3D scenarios, potentially leading to blurred anomaly boundaries, attenuated response signals, and compromised accuracy in capturing essential anomalous features.

To achieve key information focusing during the feature fusion process and enhance the network's ability to extract critical abnormal information, we embed a dual-path attention gating module in the skip connection path. This module uses the deep features from the decoder (low resolution, high semantics) as semantic guidance signals to perform spatial-channel collaborative weighting on the shallow features (high resolution, rich details) transmitted through the shortcut connection. Then, spatial attention (dilated convolution + $Sigmoid$ activation) is applied to generate a weight heatmap aligned with the resistivity mutation regions. Finally, the weighted feature map is concatenated with the up-sampled result~\cite{ref42}. The dual-path attention gating mechanism module is shown in Fig. \ref{fig2}. Here, the shallow feature map from the skip connection is denoted as $C_{shallow}$, which can be expressed as:
\begin{figure*}[!htp]
	\centering
	\includegraphics[width=5.5in]{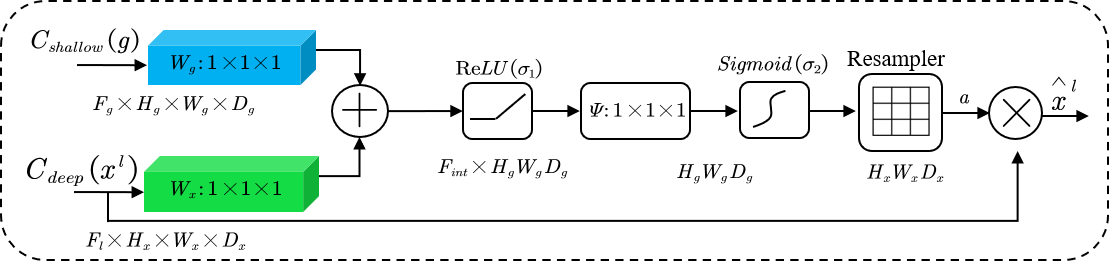}
	\caption{Diagram of the Attention gate mechanism module.}
	\label{fig2}
\end{figure*}
\vspace{-0.1in}
\begin{align}
	C_{shallow}(X, Y, Z, l)=\sum_{i=1}^{m} \sum_{j=1}^{m} \sum_{k=1}^{m} \sum_{c=1}^{L}[f(i, j, k, c) *
	\\
	\notag
	 I(S_x+i-P, S_y+j-P, S_z+k-P, c)],
\end{align}
where, $f(i, j, k, c)$ is the convolution kernel, $I$ is the feature map, and $P$ and $S$ represent the padding and stride used in the convolution. The deep feature map $C_{deep} $ from the previous transposed convolution result can be represented as:
\begin{align}
	C_{deep}(X, Y, Z, l)=\sum_{i=1}^{m} \sum_{j=1}^{m} \sum_{k=1}^{m} \sum_{c=1}^{L}[f(i, j, k, c) *
	\\
	\notag
	I(x+i-P, y+j-P, z+k-P, c)],
\end{align}
here, $x, y$ and $z$ represent the spatial coordinates of the image $ I$. The input to each layer of the decoding part consists of $C_{shallow}$ and $C_{deep}$, which are added and merged before being passed through a convolutional layer for the up-sampling operation. The merged result is then activated by the ReLU activation function, which is defined as $ReLU(x)=max(0, x)$. Subsequently, the output undergoes a $1\times 1\times 1$ convolutional layer, also known as channel pooling, which can be represented as:
\begin{equation}
	\label{eqa4}
	\Psi(X, Y, Z, l)=\sum_{c=1}^{L}[f(1,1,1, c) \times I(x, y, z, c)],
\end{equation}
where, $f(1, 1, 1, c)$ is a convolutional kernel of size $1\times 1\times 1$, which is used to apply a weighting process to each channel. This is followed by a sigmoid activation function, defined as $sigmoid(x)=1/(1+e^{-x})$, to obtain an importance score that ranges between 0 and 1. This importance score serves as a weight, which is then assigned to different parts of the feature map. The result is then multiplied with the input from the skip connections, generating the final output of the attention gate module~\cite{ref43}. As the network is trained, the attention gate weights are continuously optimized, allowing each layer to gradually achieve more accurate weight distributions. In this way, the network dynamically adjusts the weights at different levels based on the learned features, enabling more precise feature extraction and anomaly region identification.

\subsection{Loss Function}
Since the geoelectric forward response data is continuous real-valued, it is considered a regression problem. Therefore, we use mean squared error (MSE) as the loss function for the network, which measures the difference between the actual forward response data and the forward response data predicted by the network through forward propagation. It can be defined as:
\begin{equation}
\label{eqa16}
MSE=\frac{1}{N} \sum_{i=1}^N{ \left(d_i-\hat{d}_i \right) ^2, i=1, 2, 3, \cdots, N.}
\end{equation} 
Here, $N$ represents the total number of data points, while $d_i$ and $\hat{d}_i$ refer to the true and predicted forward response data, respectively. During training, the model's performance is evaluated using the loss value, and the parameters are updated via backpropagation to minimize the loss, improving the accuracy of the model's output.

\subsection{Datasets Generation}
\begin{figure}[!htp]
	\centering
	\includegraphics[width=3.5in]{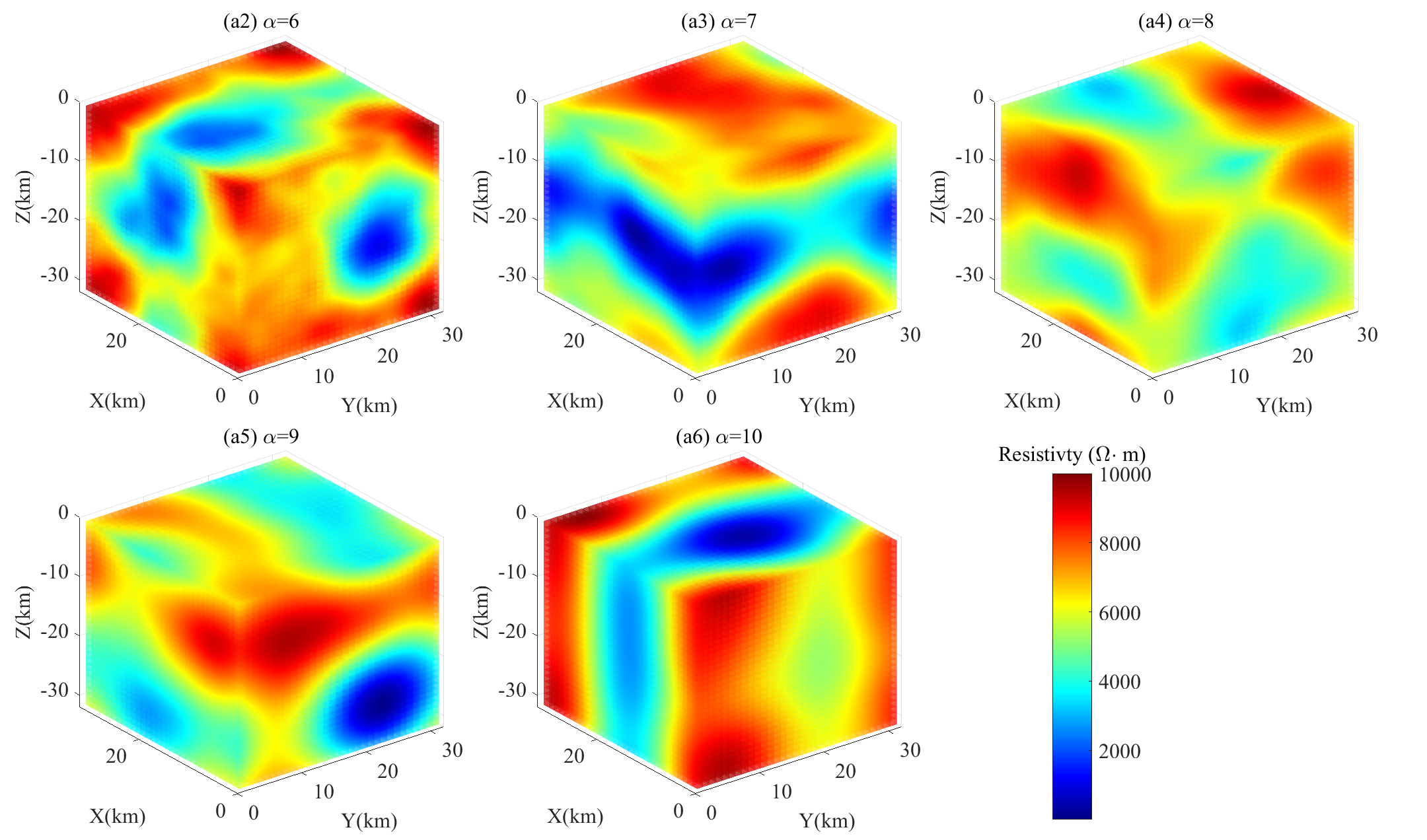}
	\caption{Examples of different conductivity models generated using Gaussian random fields (when $\alpha$ takes different values).}
	\label{fig3}
\end{figure}

Due to the high cost and complexity of field surveys, obtaining large-scale, real-world MT data is challenging. Therefore, the training and testing samples used in this study are simulated data. We discretized the subsurface study area into a $32\times 32\times 32$ 3D grid, where the X (east), Y (north), and Z (depth) directions each contain 32 grid cells, and the study area is set to a uniform grid with a cell spacing of 1 km. The 3D conductivity $\sigma (x, y, z)$ generally exhibits arbitrary spatial heterogeneity in the core domain. Thus, we employed a Gaussian random field (GRF) to generate conductivity structures controlled by the spectral method~\cite{ref31,ref44}. The spectral $P(K)$ is defined as:
\begin{equation}
	\label{eqa6}
	P (K ) \propto |K|^{-\frac{\alpha}{2}},
\end{equation} 
here, $K$ represents the wave number, and $\alpha$ represents the extension length (also referred to as the smoothing factor), with conductivity values ranging from 1 $\varOmega m$ to 10,000 $\varOmega m$. The extension length (smoothness) of the resistivity structure increases with the increase with $\alpha$, as shown in Fig. \ref{fig3}. The term "extension" refers to the spatial distribution and range of the resistivity structure, indicating the geological distribution of these underground features \cite{ref30}. We selected $\alpha$ = [6, 7, 8, 9, 10] to generate the datasets, which represent five different amplitudes of average mixtures to ensure that the generated data simulates various types of resistivity structures as accurately as possible.

During the forward modeling process, the original study area of $32\times 32\times 32$ is extended in four directions by 5 scaling grids, with a scaling factor of 1.25, resulting in an overall computational grid of $42\times 42\times 42$. Then, we use the finite element method in the subsection \ref{subA} to perform forward modeling on the simulated conductivity model. The air layer is set to the top 5 layers of the surface, and the frequency range is from $10^{-3}Hz$ to $10^{3}Hz$ in the logarithmic domain, which is divided into 16 frequency points at equal intervals.

\begin{figure}[!htp]
	\centering
	\includegraphics[width=3.5in]{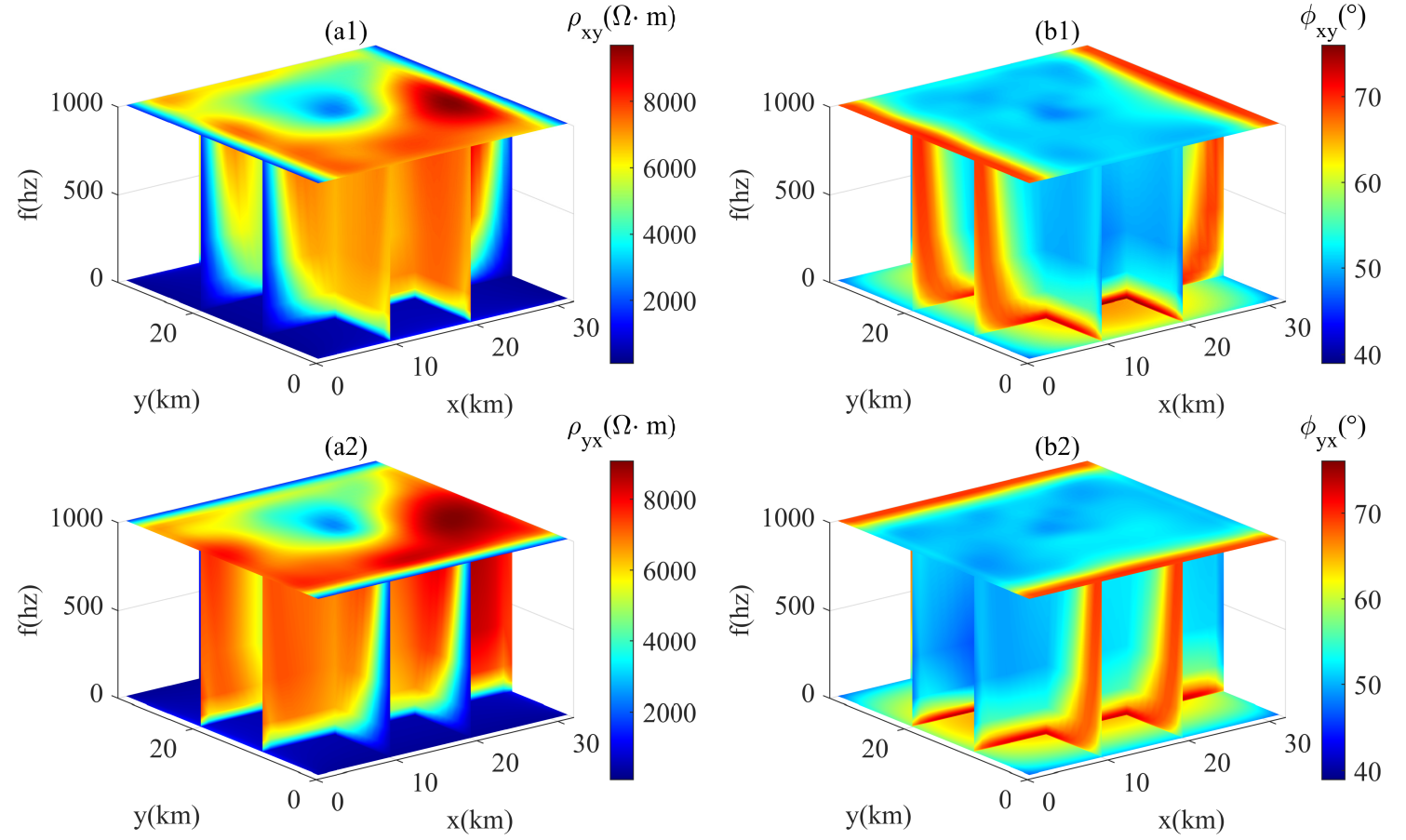}
	\caption{Forward modeling response data of the example resistivity model (at the Fig. \ref{fig3}(a1)). (a1) and (a2) represent the apparent resistivity in the XY and YX directions, respectively, while (b1) and (b2) represent the phase in the XY and YX directions, respectively.}
	\label{fig4}
\end{figure}

Based on the above strategy, we performed forward modeling using MATLAB 2024a software on a workstation equipped with an i9-14900K processor, with 8 CPU cores running simultaneously. Each 3D model's forward calculation takes approximately 2 minutes, and the entire calculation process lasted around 14 days. In total, 10,000 model samples with different distribution characteristics (2,000 for each $\alpha$ value) were generated. Fig. \ref{fig4} shows the forward response data for the model example from Fig .\ref{fig3}(a1), each sample includes a 3D resistivity model and forward response data (including apparent resistivity $\rho$ and phase $\phi$ in the XY and YX directions). Compared to previous studies, the resistivity model we constructed integrates geological factors such as geometric shape (horizontal extent, depth distribution) and spatial correlation through a Gaussian random field, which allows it to more accurately reflect actual field magnetotelluric observation data, demonstrating higher similarity and representativeness.

\subsection{Training Strategy}
We randomly split the dataset into training (85\%, 8500 samples), validation (10\%, 1000 samples), and test (5\%, 500 samples) sets. Additionally, to accelerate the network training convergence, we apply normalization to the data using formulas (\ref{eqa7}) and (\ref{eqa8}), ensuring that the input and output values of the network model are within the range of [0,1]. This can be defined as:
\begin{equation}
\label{eqa7}
{{\tilde{d}}_{i}}=\frac{{{\log }_{10}}({{d}_{i}})}{\underset{j=1,...,{{N}_{d}}}{\mathop{\max }}\,({{\log }_{10}}({{d}_{j}}))},  i=1,\cdots ,{{N}_{d}},
\end{equation}
\begin{equation}
\label{eqa8}
{{\tilde{m}}_{j}}=\frac{{{\log }_{10}}({{m}_{j}})}{\underset{k=1,...,{{N}_{m}}}{\mathop{\max }}\,({{\log }_{10}}({{m}_{k}}))},  j=1,\cdots ,{{N}_{m}}.
\end{equation}
Here, $N_d$ denotes the total number of input samples in the dataset, $\tilde{d}_i$ represents the  $i$-th normalized observation, $N_m$denotes the total number of model parameters, and $\tilde{m}_j$ represents the $j$-th normalized model parameter.

The neural network model in this study was trained using MATLAB 2024a software, with the Max Epochs set to 200, Mini batch Size set to 8, and the optimizer set to Adam. We employed a dynamic learning rate adjustment scheme with an initial learning rate of 0.001. Every 20 epochs, we multiplied the learning rate by a decay factor of 0.6. The specific training parameters of the network are shown in Table \ref{tab1}.

\section{Experiments and Analysis}\label{III}

\subsection{Evaluation Indicator}
We define two evaluation metrics, Root Mean Squared Error (RMSE) and Structural Similarity Index (SSIM), to quantitatively compare the network's prediction performance. RMSE mainly reveals relatively large data errors, and it is defined as:
\begin{equation}
	\label{eqa21}
R M S E=\sqrt{\frac{1}{N} \sum_{i=1}^{N}\left(d_{i}-\hat{d}_{i}\right)^{2}},
\end{equation}
here, $d_i$ and $\hat{d}_i$ represent the true and predicted forward response data, respectively. Different from RMSE, SSIM measures the distortion and similarity of the response data images. It is a perceptual model that aligns more closely with human visual perception \cite{ref35,ref45}. The value of SSIM ranges from [0, 1], where a value closer to 1 indicates higher similarity between the two images and better prediction performance. It is defined as:
\begin{equation}
	\label{eqa21}
	\operatorname{SSIM}(x, y)=\frac{\left(2 \mu_{x} u_{y}+C_{1}\right) \times\left(2 \sigma_{x y}+C_{2}\right)}{\left(\mu_{x}^{2}+\mu_{y}^{2}+C_{1}\right) \times\left(\sigma_{x}^{2}+\sigma_{y}^{2}+C_{2}\right)},
\end{equation}
here, $x$ and $y$ represent the true and predicted response images, respectively, $\mu_x$ and $\mu_y$ represent the mean brightness of the images, $\sigma_x$ and $\sigma_y$ represent the standard deviations of the images, and $\sigma_{xy}$ denotes the covariance. $C_1=k_1 \cdot L$ and $C_2=k_2 \cdot L$ are constants used to stabilize the computation and avoid division by zero, while $L$ represents the dynamic range. $k_1$ and $k_2$ are parameters used to adjust the contrast and brightness, typically set to 0.01 and 0.03, respectively.
\begin{table*}[h]
	\centering
	\caption{MTAG-Net model training hyperparameter settings table.}
	\label{tab1}
	\setlength{\tabcolsep}{15px}
	\begin{tabular}{cc|cc}
		\toprule
		\textbf{Hyper-parameter} & \textbf{Value} & \textbf{Hyper-parameter} & \textbf{Value} \\
		\midrule
		Max Epochs & 200 & L2 Regularization & 0.0005 \\
		Mini batch Size & 8 & Shuffle & every-epoch \\
		Optimizer & Adam & Initial Learn Rate & 0.001 \\
		Validation Frequency & 50 & Learn Rate Schedule & piecewise \\
		Gradient Threshold Method & 12norm & Learn Rate Drop Period & 20 \\
		Gradient Threshold & 0.05 & Learn Rate Drop Factor & 0.6 \\
		\bottomrule
	\end{tabular}
\end{table*}

\subsection{Network Structure Comparison}
\begin{figure}[!htp]
\centering
\includegraphics[width=3.5in]{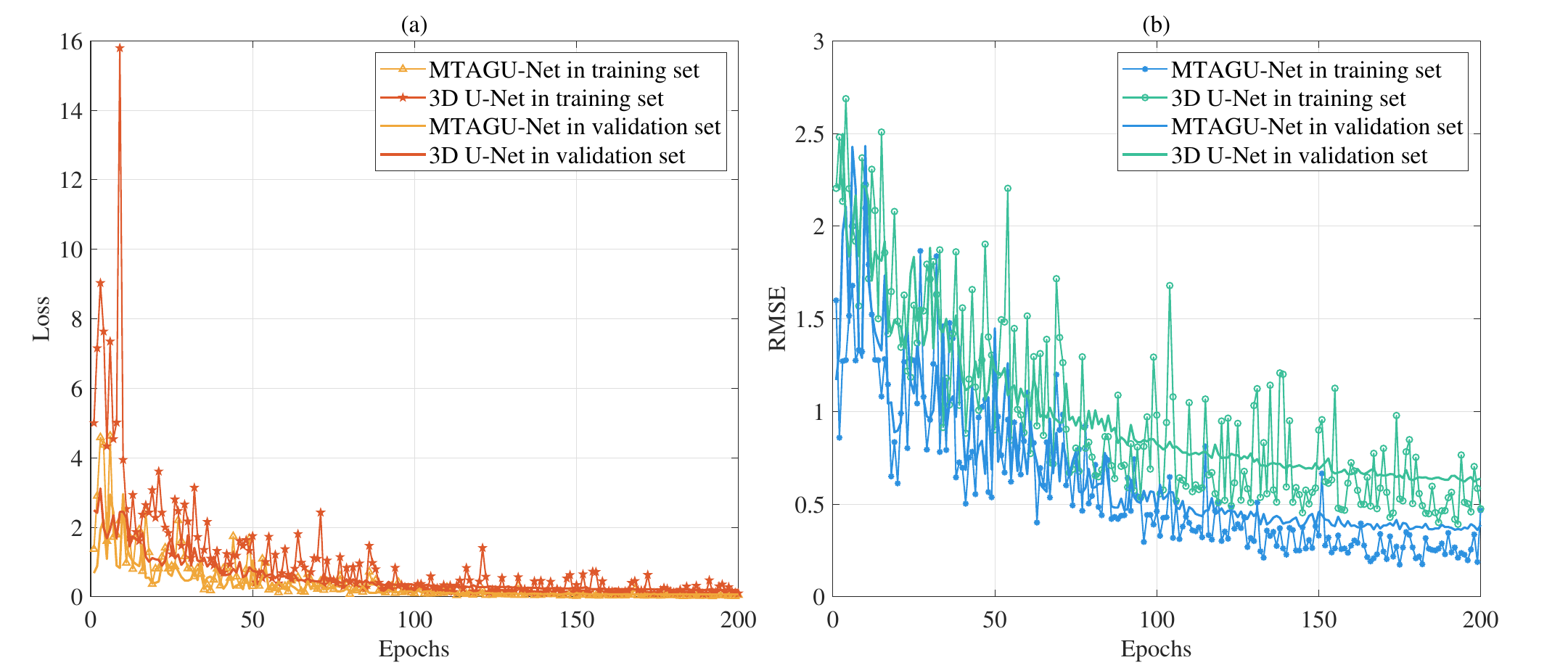}
\caption{Iteration curves of network training Loss and RMSE error.}
\label{fig5}
\end{figure}

To evaluate the forward prediction performance of the MTAGU-Net network structure, we conducted a comparative experiment with a conventional 3D U-Net network structure. To ensure fairness in the experiment, both networks were trained using the same hyperparameters (specific hyperparameter settings are shown in Table \ref{tab1}). Fig. \ref{fig5} illustrates the Loss and RMSE error reduction curves for each network model during the training process. As seen in the figure, the training performance of the MTAGU-Net model we constructed outperforms the conventional 3D U-Net model, demonstrating more stable convergence. 

\begin{figure*}[!htp]
	\centering
	\includegraphics[width=6in]{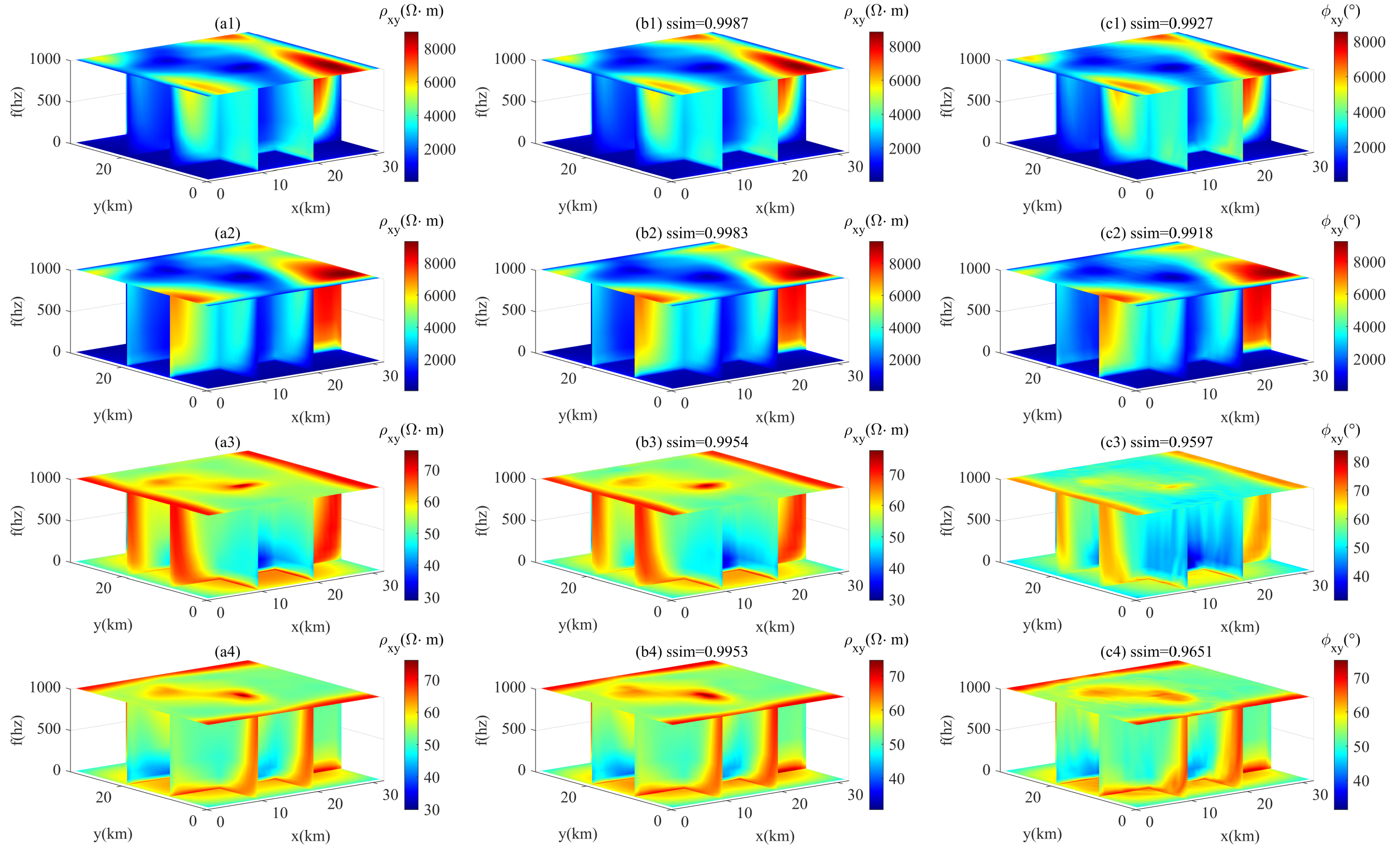}
	\caption{Comparison of network forward response data predictions. (a1)–(a4) represent the true response data (viewed resistivity and phase data in the XY and YX directions, respectively), (b1)–(b4) represent the predictions made by the MTAGU-Net network, and (c1)–(c4) show the predictions made by the 3D U-Net network.}
	\label{fig6}
\end{figure*}
\begin{figure*}[!htp]
	\centering
	\includegraphics[width=6.5in]{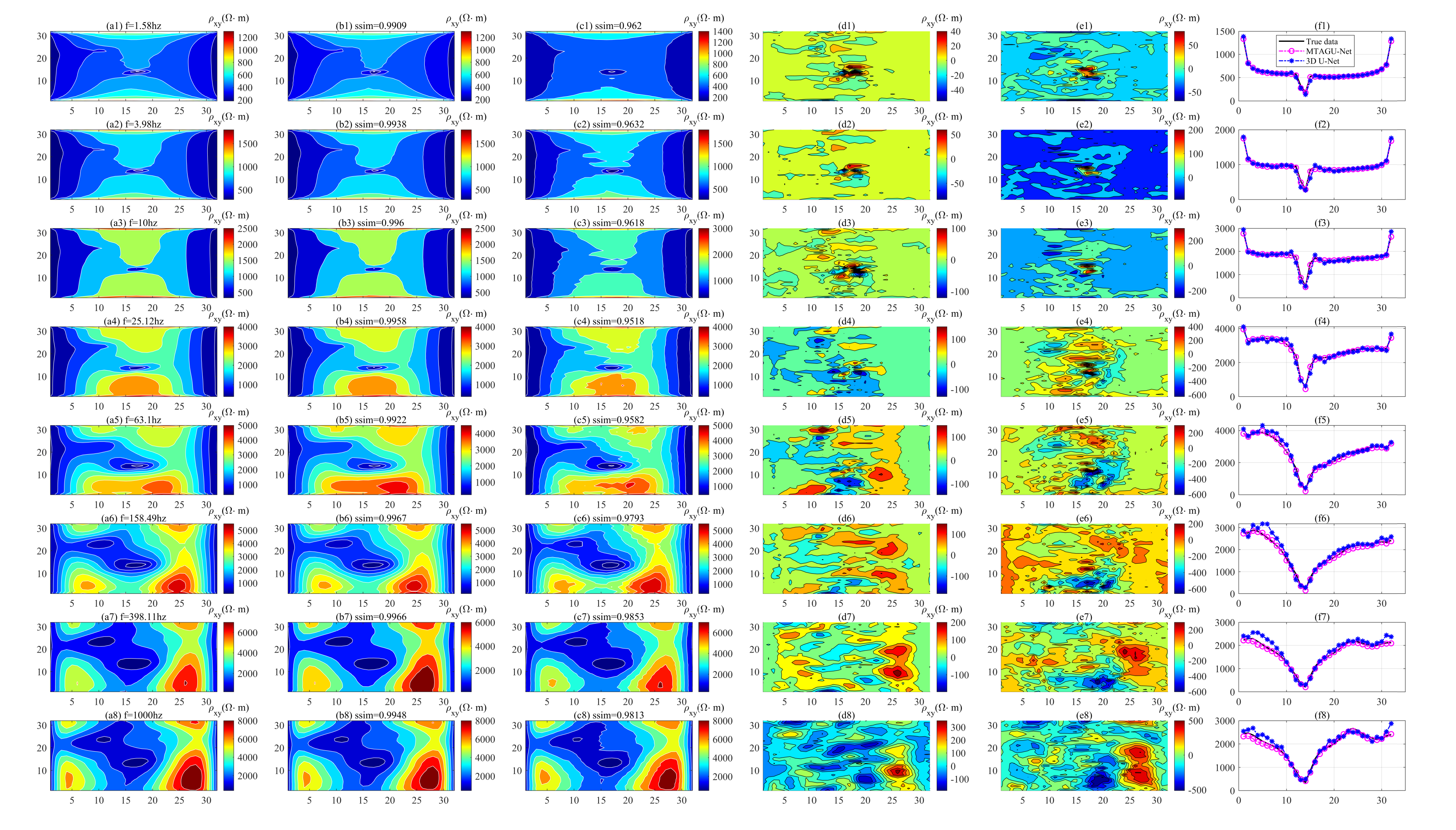}
	\caption{Slice plot of the frequency points in the XY direction resistivity from Fig. \ref{fig6}(a1). The first column shows the true response data, the second and third columns display the predictions from the MTAGU-Net and 3D U-Net networks, respectively. The fourth and fifth columns present the residuals of the predictions from the two networks, and sixth column shows the tangent fitting plot at y = 16 km.}
	\label{fig7}
\end{figure*}

To further evaluate the prediction performance of each network, we randomly selected a resistivity model from the test dataset for analysis. As shown in Fig. \ref{fig6}, overall, both network models are able to accurately predict the basic distribution characteristics of the model response data, but there are certain differences in local details. To analyze the prediction performance of each network in local details, we sliced the XY-direction apparent resistivity in Fig. \ref{fig6}(a1) at different frequencies, selecting eight frequency points [1.58 Hz, 3.98 Hz, 10 Hz, 25.12 Hz, 63.1 Hz, 158.49 Hz, 398.11 Hz, 1000 Hz], and plotted the 2D cross-sectional maps. As shown in Fig. \ref{fig7}, the prediction results of MTAGU-Net are highly consistent with the actual response data, with SSIM values maintained above 0.99, and the residuals range from [-100 $\varOmega m$ , 300 $\varOmega m$]. In contrast, the prediction results of the 3D U-Net show significant deviations, with abnormal shape deformations and rough boundaries, and the residuals are generally distributed between [-600 $\varOmega m$, 500 $\varOmega m$]. Furthermore, from the fitting tangent plots in the sixth column, it is visually clear that the trend of MTAGU-Net (blue dashed line) matches the actual data (black solid line) closely, with the curves aligning well. Although the trend of the 3D U-Net curve (magenta solid line) is generally consistent, there are still significant deviations in local areas, particularly in Fig. \ref{fig7}(f6) to Fig. \ref{fig7}(f8), where the curve shows noticeable deviations.
In addition, we calculated the average SSIM and RMSE metrics of the network's prediction results across 500 test set models, with the summarized results presented in Table \ref{tab2}. It can be observed that, in terms of both SSIM and RMSE metrics, the MTAGU-Net outperforms the conventional 3D U-Net network. It should be added, due to the addition of the attention gate mechanism, the MTAGU-Net network structure has more layers compared to the conventional 3D U-Net, leading to a longer training time of approximately 21 hours, which is about 5 hours longer than the 16 hours required by the 3D U-Net.
\begin{table}[h]
	\centering
	\caption{Summary of evaluation metrics for forward response data prediction of each network.}
	\label{tab2}
	
	\begin{tabular}{ccccc}
		\toprule
		\textbf{Network} & \textbf{Train time} & \textbf{Response data} & \textbf{RMSE} & \textbf{SSIM} \\
		\midrule
		MTAGU-Net & 1261 min & \(\rho_{xy}\) & 0.2661 & 0.9923 \\
		& & \(\rho_{yx}\) & 0.2693 & 0.9934 \\
		& & \(\phi_{xy}\) & 0.2471 & 0.9921 \\
		& & \(\phi_{yx}\) & 0.2362 & 0.9923 \\
		\midrule
		\multicolumn{1}{c}{3D U-Net} & \multicolumn{1}{c}{983 min}  & \(\rho_{xy}\) & 0.3704 & 0.9667 \\
		& & \(\rho_{yx}\) & 0.3787 & 0.9661 \\
		& & \(\phi_{xy}\) & 0.3691 & 0.9676 \\
		& & \(\phi_{yx}\) & 0.3603 & 0.9649 \\
		\bottomrule
	\end{tabular}
\end{table}

\begin{figure}[!htp]
	\centering
	\includegraphics[width=3.5in]{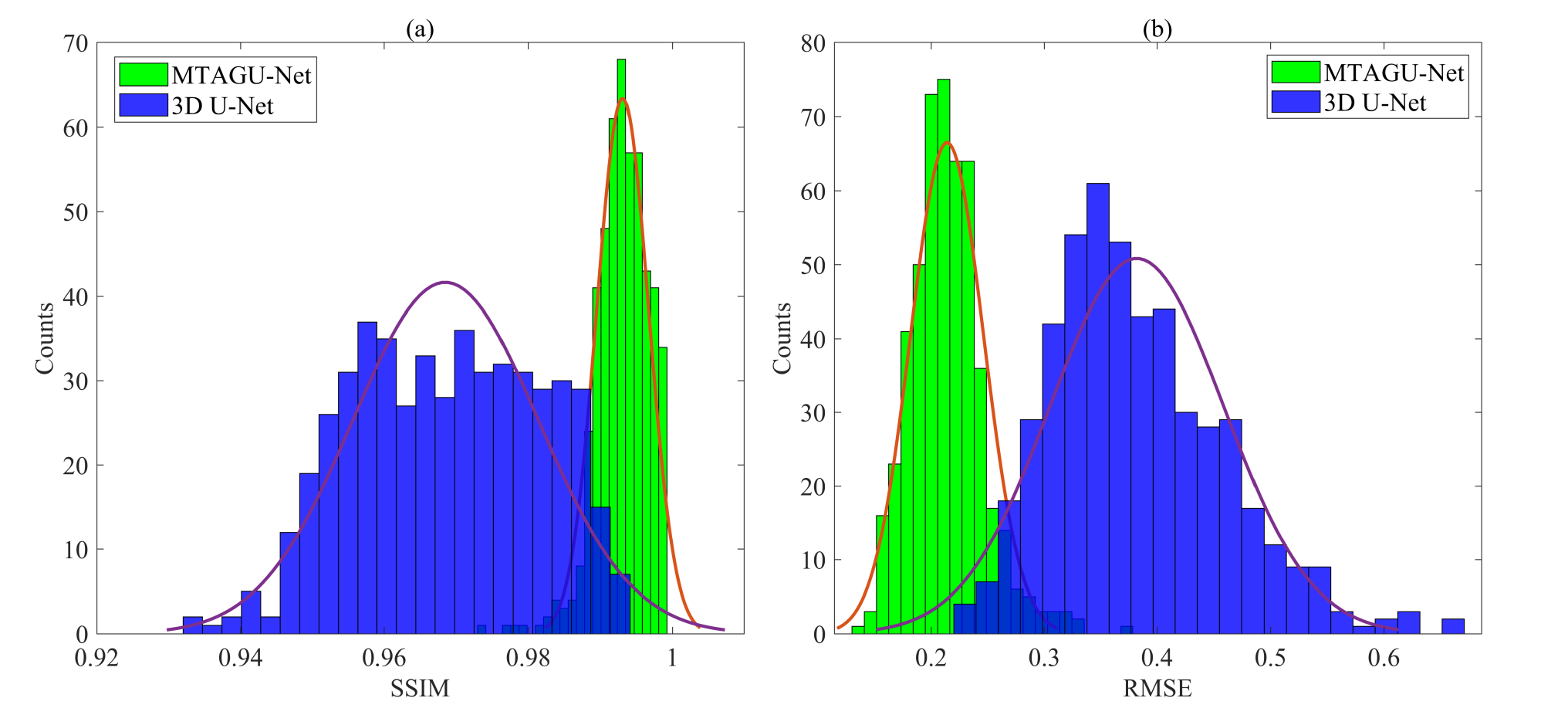}
	\caption{Distribution map of each network prediction evaluation index. (a) is the SSIM metric, (b) is the RMSE metric.}
	\label{fig8}
\end{figure}

To analyze the overall distribution of the metrics, we also calculated the distribution histograms of the average SSIM and RMSE metrics for the four response datasets. As shown in Fig. \ref{fig8}, for the SSIM metric, the predictions from MTAGU-Net consistently remained above 0.98, indicating a high similarity to the true response data, whereas the SSIM values for 3D U-Net were distributed between 0.93 and 0.99, showing significant deviation. Similarly, for the RMSE metric, the error distribution for MTAGU-Net was noticeably smaller than that of 3D U-Net, indicating higher prediction accuracy. Based on the above analysis, we can conclude that although the inclusion of the attention gate mechanism makes the network structure more complex and increases the training time, MTAGU-Net significantly outperforms the conventional 3D U-Net in terms of training convergence and prediction accuracy. Therefore, this investment is worthwhile, as MTAGU-Net provides more accurate results in forward prediction tasks.

\subsection{Datasets Capacity Comparison}
\begin{figure}[!htbp]
	\centering
\includegraphics[width=3.5in]{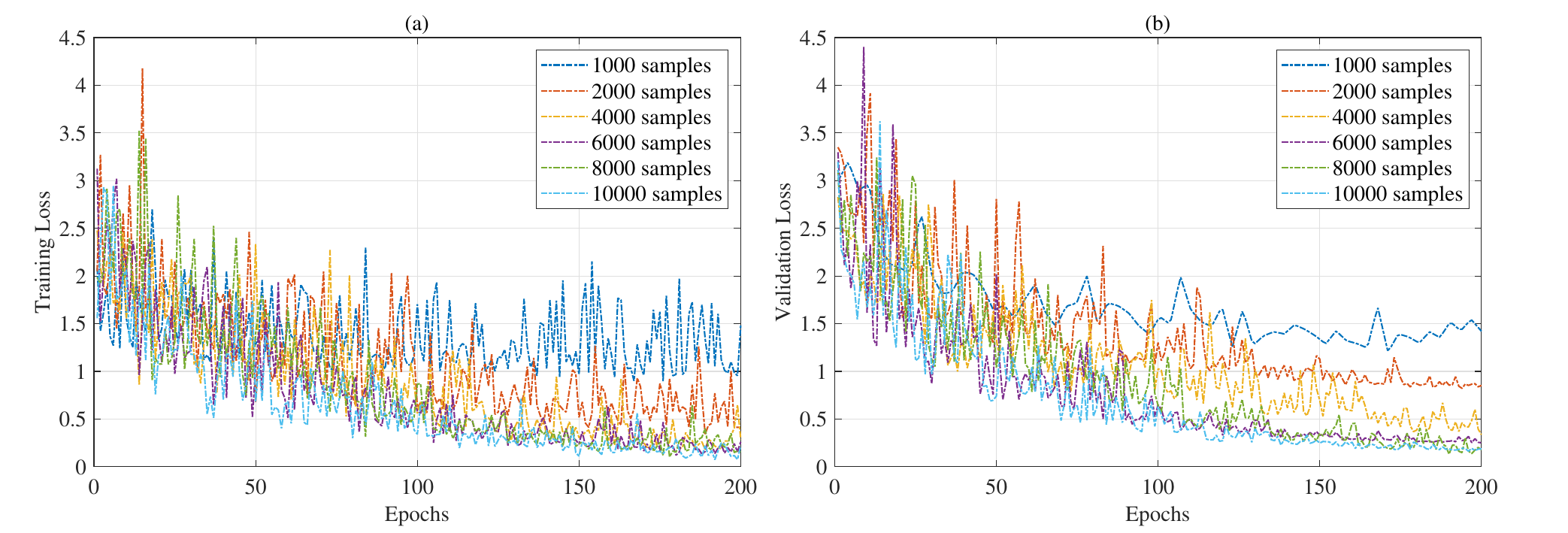}
	\caption{Network training loss curves for different dataset sizes, evaluated on the training and validation sets. (a) is the training loss; (b) is the validation loss.}
	\label{fig9}
\end{figure}

The size of the training dataset has a significant impact on the training and prediction performance of the network model. When the dataset is too small, the model may get stuck in a local optimum, making it difficult for the Loss value to converge during training. On the other hand, when the dataset is too large, it requires more computational resources, leading to a substantial increase in training time \cite{ref30}. Therefore, in an ideal scenario, the dataset should be large enough to ensure that the model can learn sufficient features from the data, but not so large as to waste computational resources and time. To analyze the impact of training sample size on network training time and test error, we trained the network using [1000, 2000, 4000, 6000, 8000, 10000] samples, and compared the effect of different dataset sizes on network performance. Fig. \ref{fig9} shows the loss curves of the training and validation sets for different sample sizes. It can be observed that when the number of training samples is below 6000, the Loss curve fluctuates significantly, and the network struggles to converge with this sample size. However, when the sample size exceeds 6000, and with increasing sample size, the network training process gradually stabilizes, and the Loss value converges to a lower level.

\begin{figure*}[!htbp]
	\centering
	\includegraphics[width=6in]{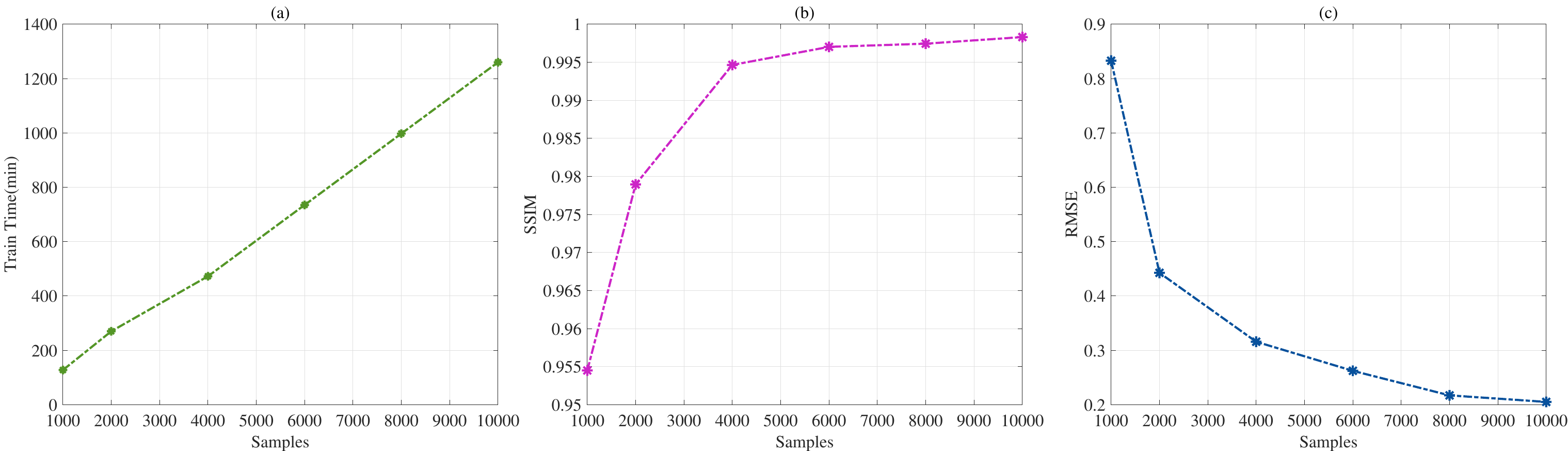}
	\caption{Relationship between dataset size, training time, and test set error. (a) is the training time, (b) is the SSIM metric, and (c) is the RMSE metric.}
	\label{fig10}
\end{figure*}

Fig. \ref{fig10} illustrates the relationship between dataset size and various metrics (training time, SSIM, RMSE). From Fig. \ref{fig10}(a), it can be observed that the training time increases linearly with the number of training samples. It is worth noting that, in Fig. \ref{fig10}(b) for SSIM and Fig. \ref{fig10}(c) for RMSE, it is evident that when the dataset size is smaller than 6000, the variations in SSIM and RMSE are relatively large. However, when the dataset size exceeds 6000, the rate of change in SSIM and RMSE gradually slows down. According to the elbow method, this suggests that the network achieves good predictive accuracy with 6000 training samples, and the training time is approximately two-thirds of that required for 10,000 samples. Therefore, within a certain range, the number of training samples has a significant impact on network performance. However, beyond a certain threshold, further increasing the sample size has a diminishing effect on improving test error. In practical applications, an appropriate dataset size can be selected based on specific accuracy requirements.
\begin{figure*}[!htbp]
	\centering
	\includegraphics[width=6in]{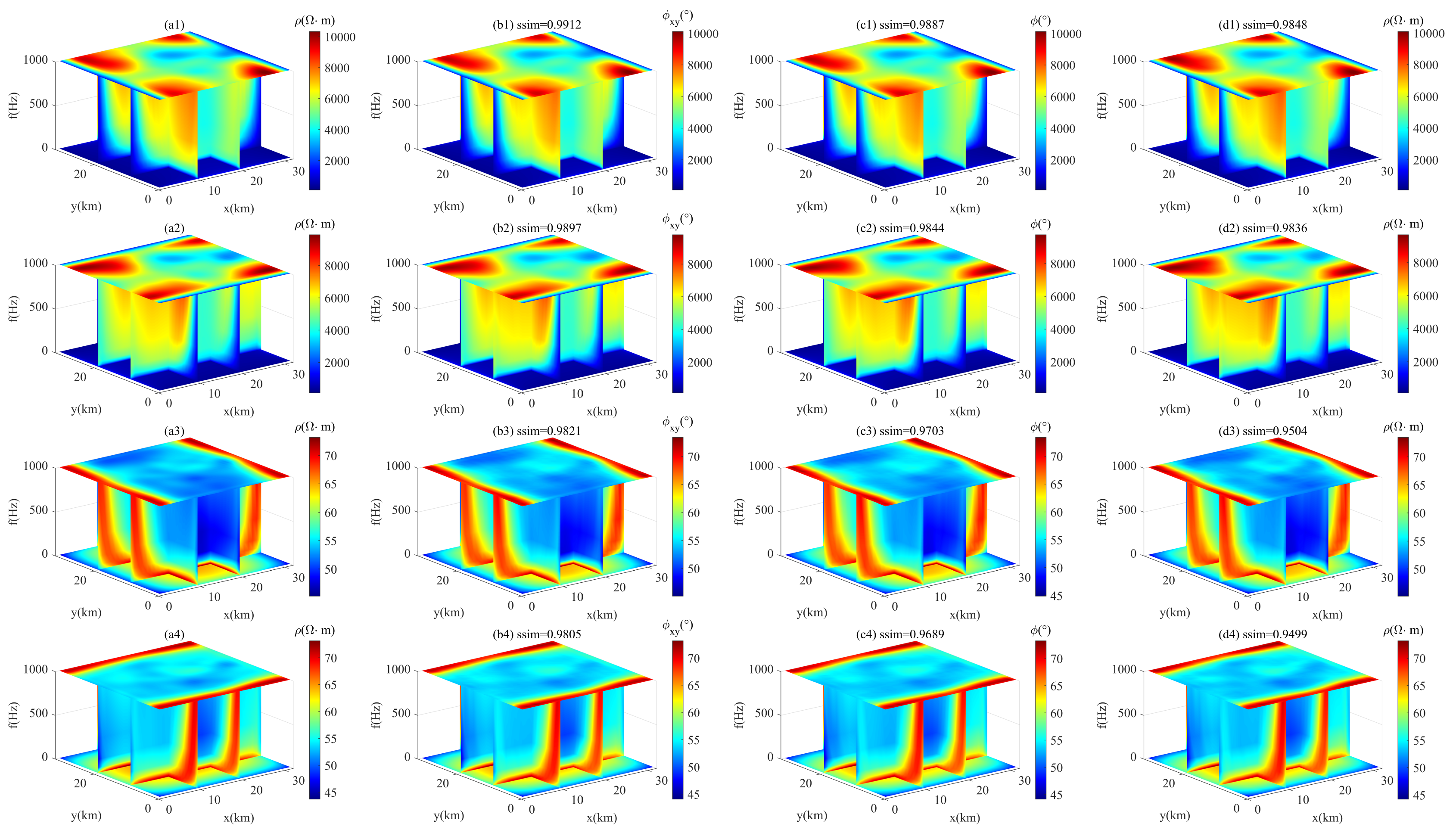}
	\caption{MTAGU-Net prediction results under different noise levels. The first column shows the true response data, while columns 2 to 4 show the model-predicted response data with noise levels of 3\%, 5\%, and 10\%, respectively.}
	\label{fig11}
\end{figure*}
\subsection{Stability Analysis}
To evaluate the stability of the model's predictions, we added Gaussian random noise at levels of 3\%, 5\%, and 10\% to the input data and analyzed the model's performance under different noise levels. As shown in Fig. \ref{fig11}, overall, the predicted response data across different noise levels appear similar, but there are some differences in the local regions.

\begin{figure*}[!htbp]
	\centering
	\includegraphics[width=6.5in]{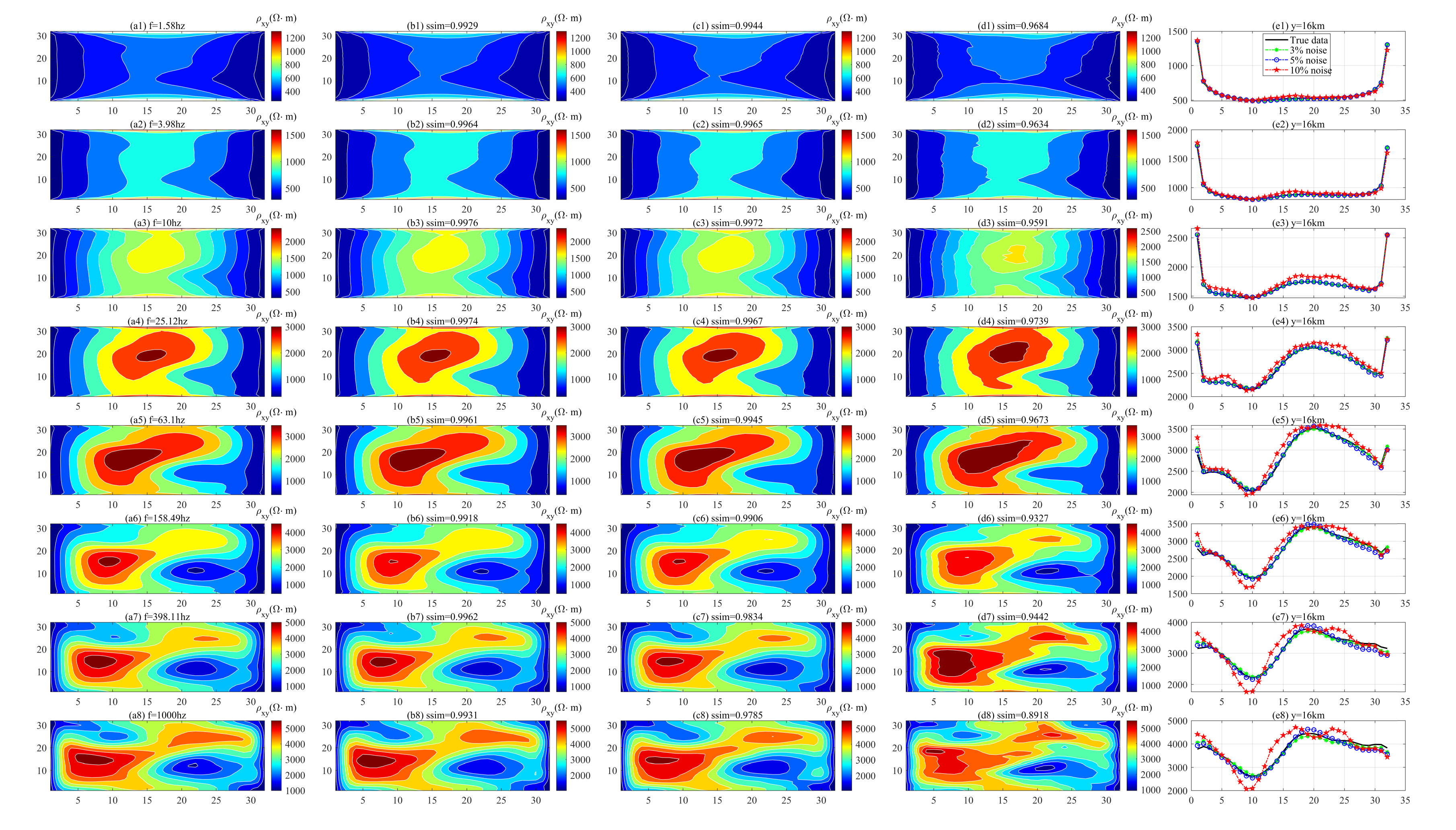}
	\caption{Frequency slice plot of apparent resistivity in the XY direction (from the Fig. \ref{fig11}(a1)). The first column shows the true response data, the second to fourth columns display the predicted results with 3\%, 5\%, and 10\% noise levels in the input data, respectively, and the fifth column shows the tangent fitting plot at y = 16 km.}
	\label{fig12}
\end{figure*}
\begin{figure}[!htbp]
	\centering
	\includegraphics[width=3.5in]{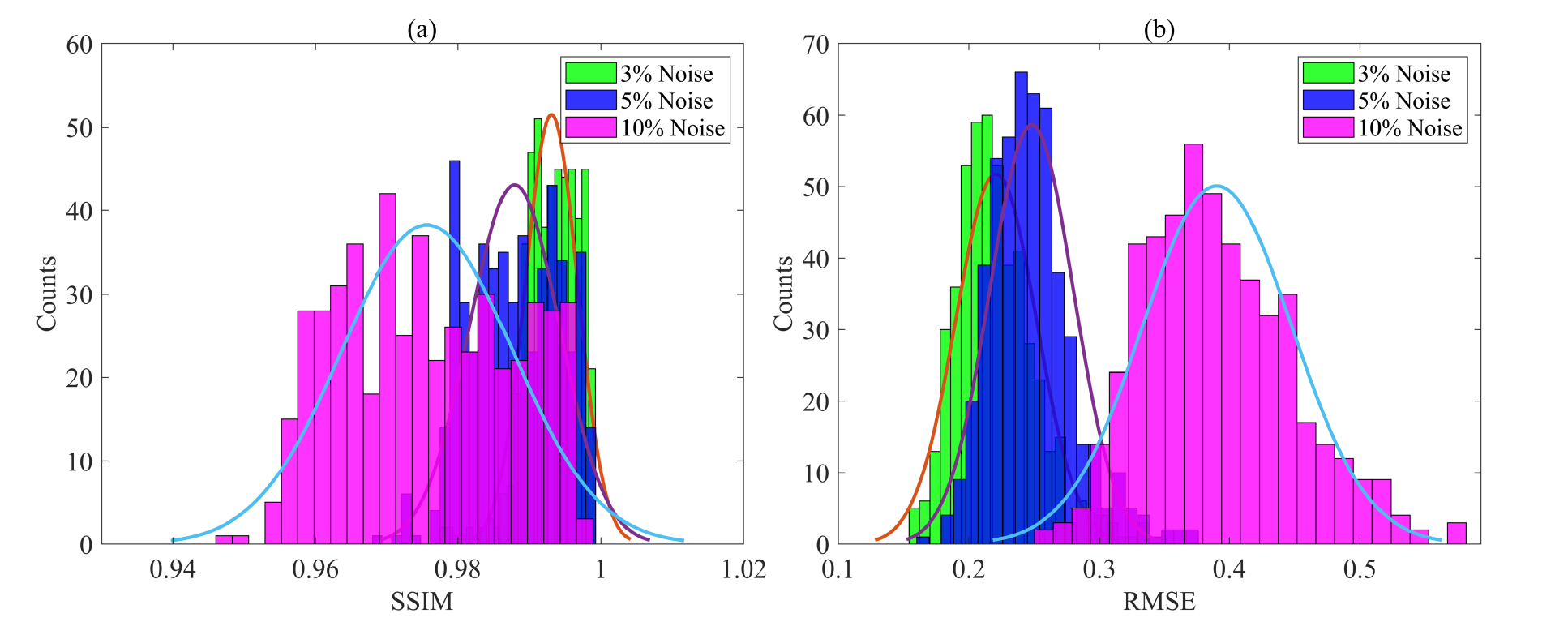}
	\caption{Distribution of prediction and evaluation indexes of different noise levels. (a) is the SSIM metric, (b) is the RMSE metric.}
	\label{fig13}
\end{figure}
To analyze the prediction results in detail, we divided the frequency range of Fig. \ref{fig11}(a1) and plotted 2D contour maps. As shown in Fig. \ref{fig12}, when the input data contains 3\% and 5\% noise, although the model's predictions exhibit flaws to varying degrees, it can still accurately predict the abnormal distribution and shape of the response data. The SSIM remains above 0.97, and the data fitting curve closely matches the ground truth. However, when the noise level is 10\%, the predictions show significant deviations due to the higher amount of noise in the input data. The abnormal regions are severely distorted, with the most serious deformation appearing in Fig. \ref{fig12}(d8), where the SSIM is only 0.8918. Moreover, the fitting curve for the 10\% noise level fluctuates more significantly, particularly in Fig. \ref{fig12}(e5) to \ref{fig12}(e8), where noticeable oscillations in the curve are observed.

\begin{figure}[!htbp]
	\centering
	\includegraphics[width=3.5in]{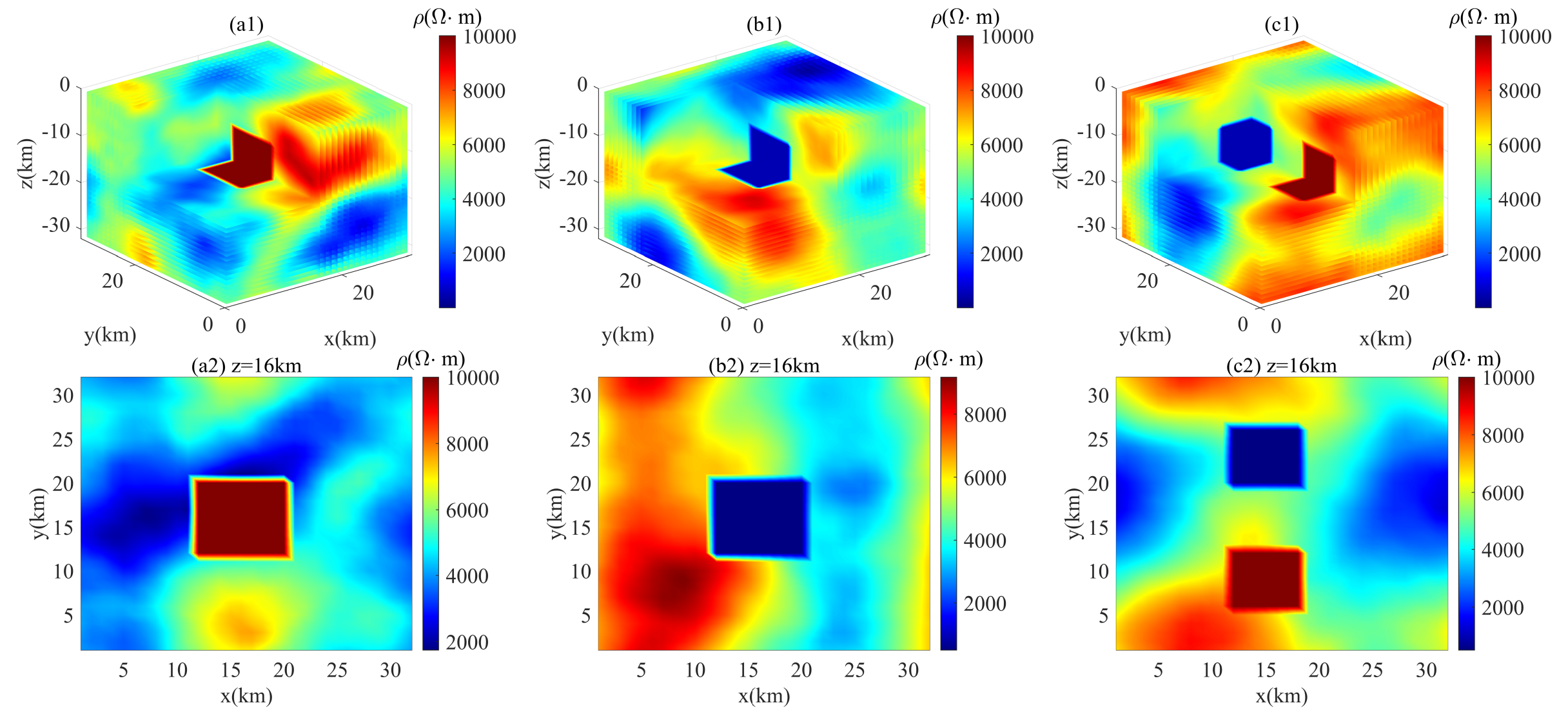}
	\caption{Embedding regular anomaly blocks into the initial complex model samples. (a1) shows a single high-resistance anomaly, (b1) shows a single low-resistance anomaly, and (c1) shows a combination of one high-resistance and one low-resistance anomaly block.}
	\label{fig14}
\end{figure}

\begin{figure*}[!htbp]
	\centering
	\includegraphics[width=5.5in]{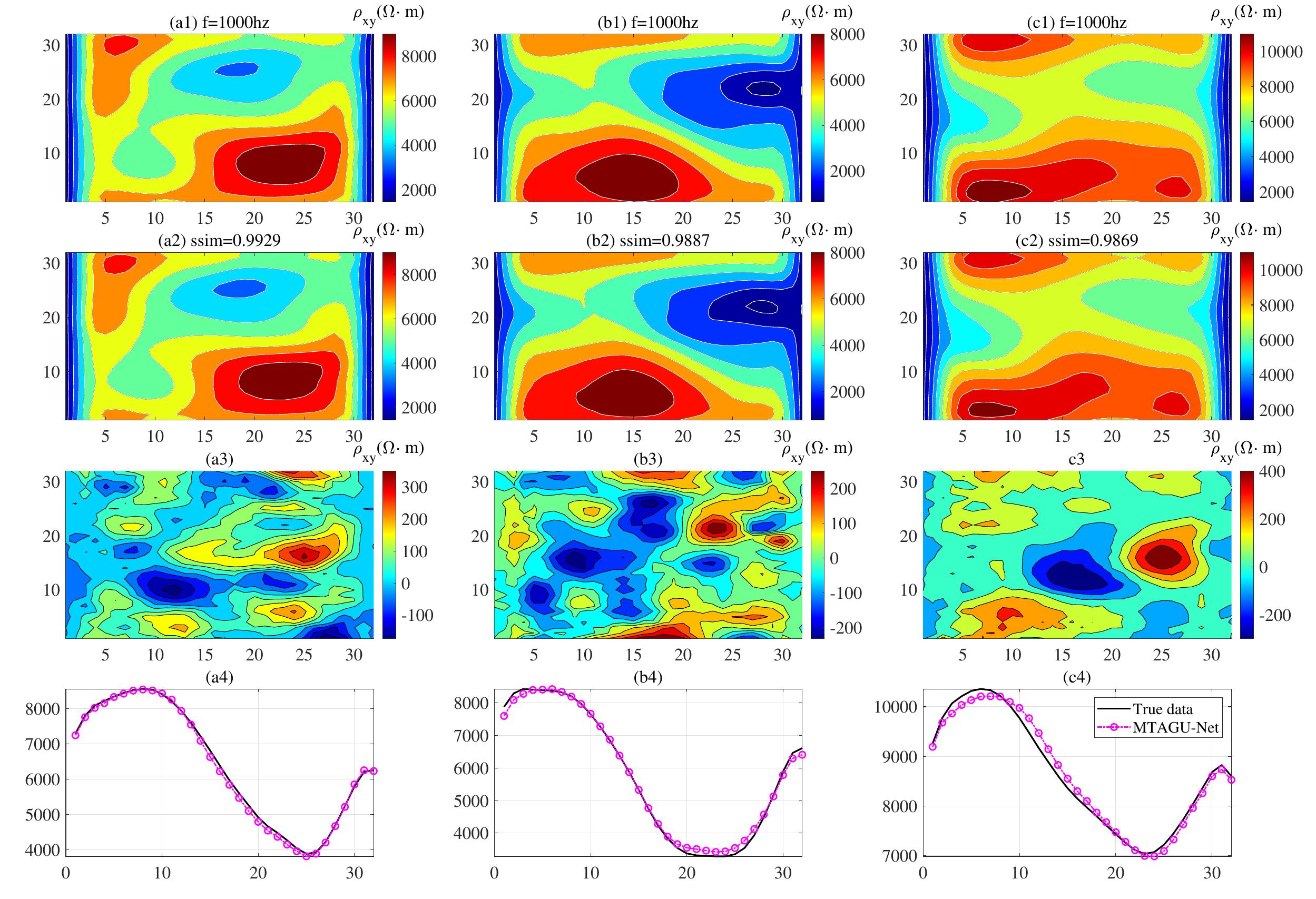}
	\caption{Comparison of the predicted response data for the discontinuous conductivity model. (a1)-(c1) show the true response data for high-resistance anomaly block, low-resistance anomaly block, and high-low combination resistance anomaly block, respectively. (a2)-(c2) display the corresponding predicted data from the MTAGU-Net. (a3)-(c3) represent the data residuals. (a4)-(c4) show the corresponding fitting curves (with the tangent point at y = 16 km).}
	\label{fig15}
\end{figure*}
The SSIM and RMSE distribution histograms at different noise levels in Fig. \ref{fig13} indicate that when the input data contains 3\% and 5\% noise, the network's predicted SSIM remains above 0.97, with the MTAGU-Net predictions showing little sensitivity to the noise. However, when the noise level reaches 10\%, the SSIM values range from 0.94 to 0.99, with larger deviations. Based on this analysis, we can conclude that when the noise level in the input data is low (0\%–5\%), the network can still effectively extract weak signals and accurately identify anomalous regions in the response data. In contrast, when the noise level increases ($\geq$ 10\%), the network's predictions are significantly affected. While the overall distribution trend can still be predicted, local deviations become much larger.

\subsection{Generalization Analysis}
In deep learning forward modeling, generalization refers to the predictive ability of a trained model on unseen input data (such as new model samples) \cite{ref46}. Therefore, a model with good generalization not only performs well on the training data but should also make accurate predictions on new, previously unseen samples. Since our theoretical model training samples are based on continuous conductivity generated by a 3D Gaussian random field, discontinuous structures may also occur in practical applications. Here, we created a new dataset that differs from the continuous conductivity structures described above. Specifically, we first used the initial model generated by the Gaussian random field as the background conductivity, and then embedded several regular anomalous resistivity blocks within the study area, as shown in Fig. \ref{fig14}. These blocks include high-resistance anomalies (10,000 $\varOmega m$), low-resistance anomalies (100 $\varOmega m$), and a combination of high and low resistance anomalies. These new datasets have never appeared in the model training data.

We used the pre-trained MTAGU-Net network to predict the forward response data for a new dataset. Fig. \ref{fig15} presents the cross-sectional map of the predicted XY-direction apparent resistivity at a frequency of 1000 Hz. It can be observed that the forward response of the three non-continuous resistivity models is largely consistent with the actual data, with SSIM values remaining above 0.98, indicating a high degree of image similarity. Moreover, from the fitting curves, it is evident that when a single high-low resistivity anomaly block is embedded in the model, the curve fitting is quite accurate, with only minor deviations at the inflection points. However, when the model includes two anomaly blocks, the increased complexity leads to a noticeable increase in fitting error Fig. \ref{fig15}(c4) and some shifts. Overall, despite the presence of local deviations, the network is still able to accurately predict the response data for an unseen dataset, with the overall image features remaining consistent, suggesting that the network has good generalization ability.

\section{Conclusion}\label{IV}
In this paper, we propose a deep learning-based 3D magnetotelluric forward modeling method. Specifically, we propose a neural network architecture based on an attention gating mechanism (MTAGU-Net). By embedding the attention gating module into the skip connections between the encoder and decoder of the network, the model can effectively integrate key information from shallow layers when decoding deep layers feature maps, thereby enhancing its ability to extract crucial anomaly features. This design enables the network to effectively integrate key information from higher layers while decoding lower-level feature maps, thereby enhancing its ability to extract features related to key anomalies. Numerical experiments demonstrate that MTAGU-Net effectively captures the relationships between data features during training and shows better convergence and prediction accuracy compared to conventional 3D U-Net networks. Although the training process is time-consuming, once the model is trained, the network can perform forward simulations with extremely high efficiency, achieving a simulation speed thousands of times faster than traditional 3D numerical methods (for example, traditional FEM requires about 120 seconds to compute one model sample, while MTAGU-Net predicts in less than 0.01 seconds). This advantage enables MTAGU-Net to demonstrate significantly higher efficiency and promising application potential when handling large-scale datasets forward simulations.

\section*{Acknowledgments}
The authors would like to sincerely thank the editors and reviewers for their insightful and constructive feedback, greatly contributed to enhancing the quality of the manuscript.

\vspace{-0.5in}
\begin{IEEEbiography}[{\includegraphics[width=1in,height=1.25in,clip,keepaspectratio]{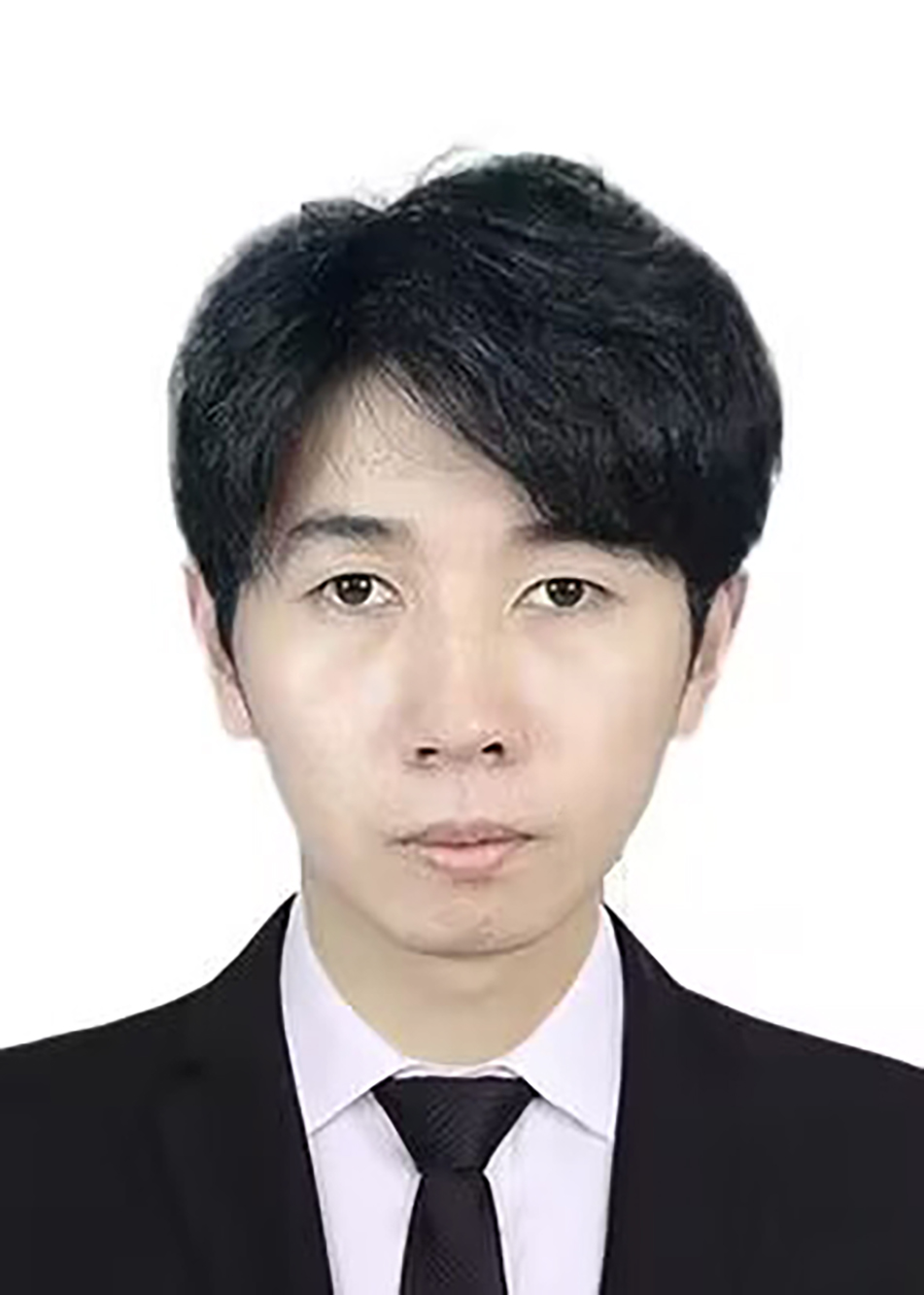}}]{Xin Zhong}
	is currently pursuing a B.E. degree in Computer Science and Technology at Jiangxi University of Science and Technology, Ganzhou, China. His research interests include: magnetotelluric data processing, numerical methods for partial differential equations, machine learning and deep learning.
\end{IEEEbiography}
\vspace{-0.5in}
\begin{IEEEbiography}[{\includegraphics[width=1in,height=1.25in,clip,keepaspectratio]{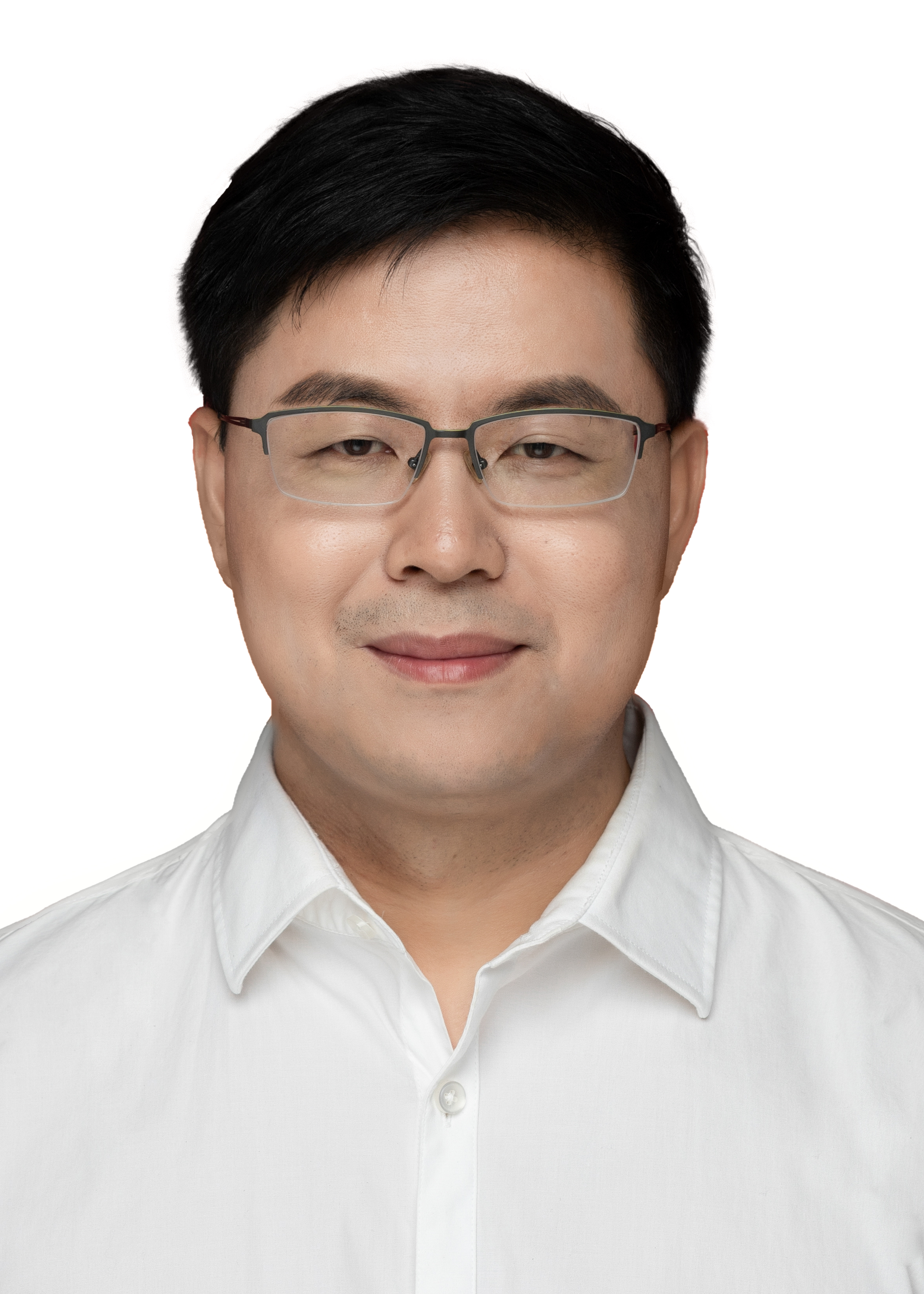}}]{Weiwei Ling}
 received the Ph.D. degree from Central South University, Changsha, China, in 2023. He is currently a Professor with Jiangxi College of Applied Technology, Ganzhou, China. His research interests include: magnetotellurics forward and inversion modeling, numerical approaches to partial differential equations, mathematical modeling and deep learning.
\end{IEEEbiography}
\vspace{-0.5in}
\begin{IEEEbiography}[{\includegraphics[width=1in,height=1.25in,clip,keepaspectratio]{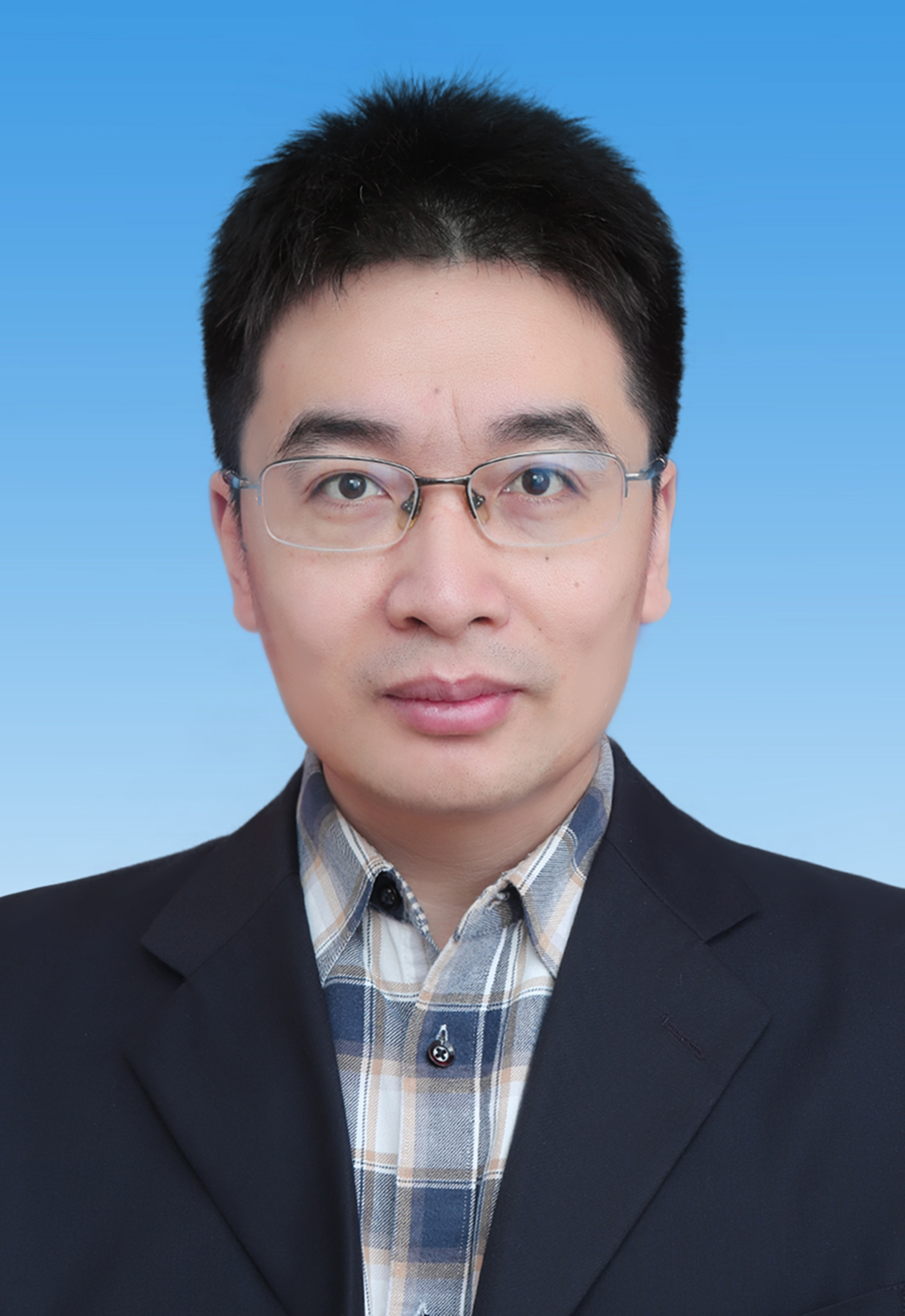}}]{Kejia Pan}
 received the Ph.D. degree from Fudan University, Shanghai, China, in 2009. He is currently a Professor with the School of Mathematics and Statistics, Central South University, Changsha, China. His research interests include forward modeling and inversion of geophysical data, especially for electromagnetic method.
\end{IEEEbiography}
\vspace{-0.5in}
\begin{IEEEbiography}[{\includegraphics[width=1in,height=1.25in,clip,keepaspectratio]{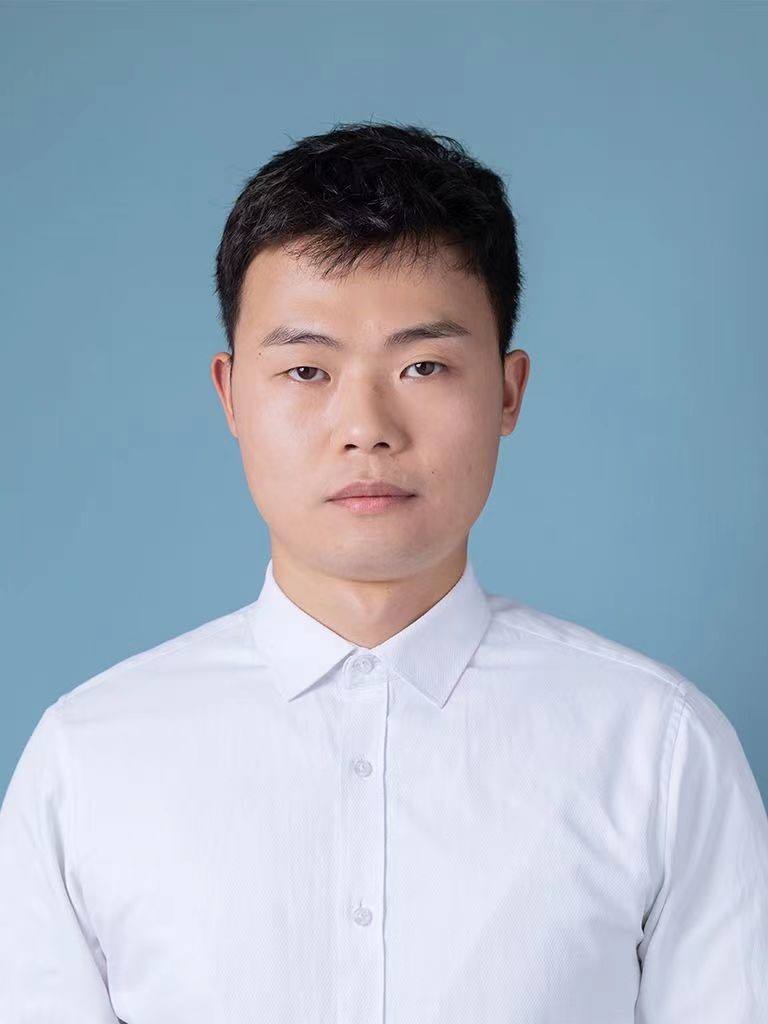}}]{Pinxia Wu}
 received the Ph.D. degree from Central South University, Changsha, China, in 2023. He is currently a Postdoc at Institute of Applied Physics and Computational Mathematics, Beijing, China. His research interests include: numerical solution of partial differential equations, mathematical modeling and machine learning.
\end{IEEEbiography}
\vspace{-0.5in}
\begin{IEEEbiography}[{\includegraphics[width=1in,height=1.25in,clip,keepaspectratio]{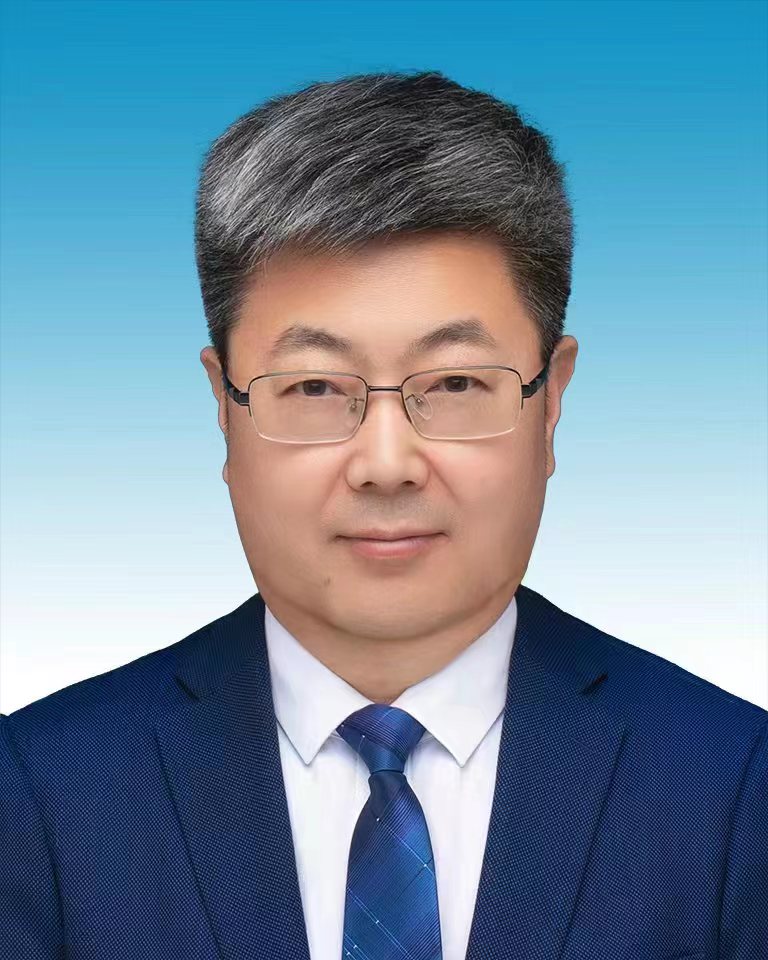}}]{Jiajing Zhang}
 received the Ph.D. degree from the Chinese Academy of Geological Sciences, Beijing, China, in 2009. He is currently a Professor with Jiangxi College of Applied Technology, Ganzhou, China. His research interests include: geological exploration, metallogenic regularity, metallogenic series and mineral exploration.
\end{IEEEbiography}
\vspace{-0.5in}
\begin{IEEEbiography}[{\includegraphics[width=1in,height=1.25in,clip,keepaspectratio]{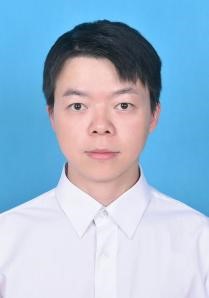}}]{Zhiliang Zhan}
	received the B.E. degree from the Jiangxi University of Science and Technology, Ganzhou, China, in 2022. He is currently pursuing a Master's degree in Computer Science and Technology at Jiangxi University of Science and Technology, Ganzhou, China. His research interests include: magnetotelluric inversion, mathematical modeling, and deep learning.
\end{IEEEbiography}

\begin{IEEEbiography}[{\includegraphics[width=1in,height=1.25in,clip,keepaspectratio]{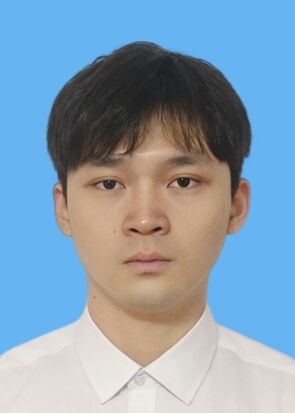}}]{Wenbo Xiao}
	received the B.E. degree from the Xinyu University, Xinyu, China, in 2022. His research interests include: numerical solution of partial differential equations, mathematical modeling and machine learning.
\end{IEEEbiography}

\end{document}